\documentclass[10pt,twocolumn,letterpaper]{article}
\usepackage[margin=0.75in,columnsep=0.25in]{geometry}
\usepackage[T1]{fontenc}
\usepackage{times}
\usepackage{cite}
\usepackage{graphicx}
\usepackage{caption}
\usepackage{url}
\usepackage{xcolor}

\usepackage[
    colorlinks=true,
    urlcolor=blue,
    linkcolor=black,
    citecolor=black
]{hyperref}
\usepackage[hyperpageref]{backref}   
\renewcommand*{\backref}[1]{}\renewcommand*{\backrefalt}[4]{\ifcase #1\or #2\else #2\fi}

\usepackage{amsmath}
\usepackage{amssymb}
\usepackage{amsfonts}
\usepackage{booktabs}
\usepackage{multirow}
\usepackage[table]{xcolor}
\usepackage{nicefrac}
\usepackage{xspace}
\usepackage{pifont}%
\usepackage{balance}%

\newcommand{\paper}{\textbf{SemCam}\xspace}

\makeatletter
\def\thebibliography#1{\section*{References}
 \list{\relax}{\setlength{\itemindent}{0pt}\setlength{\leftmargin}{1em}\itemsep 4pt plus 1pt minus 1pt\fontsize{8}{9}\selectfont}%
 \parindent\z@\parskip\z@ plus .1pt\relax
 \def\newblock{\hskip .11em plus .33em minus .07em}%
 \sloppy\clubpenalty4000\widowpenalty4000\sfcode`\.=1000\relax}

\def\@biblabel#1{[#1] \ }
\makeatother

\title{SemCam: Semantic Camera Motion Control for Video Generation}

\newcommand{\videopage}{https://semcam.github.io}

\author{%
Janna Bruner$^{1,2*}$ \quad
Omer Talmi$^{2*}$ \quad
Ianir Ideses$^{1}$ \quad
Lior Fritz$^{1}$ \quad
Lior Wolf$^{2}$ \quad
Sagie Benaim$^{1,3}$\\[2pt]
{\normalsize
$^{1}$Amazon Prime Video \quad
$^{2}$Tel Aviv University \quad
$^{3}$Hebrew University of Jerusalem}\\[4pt]
{\small\url{\videopage}}
}
\date{}

\begin{document}

\newcommand{\tlbl}[1]{$\vcenter{\hbox{\rotatebox{90}{{\fontfamily{augie}\selectfont\fontsize{7.5}{7.5}\selectfont #1}}}}$}
\newcommand{\tlblb}[2]{$\vcenter{\hbox{\rotatebox{90}{{\fontfamily{augie}\selectfont\fontsize{7.5}{8}\selectfont \shortstack{#1\\ #2}}}}}$}
\newcommand{\tlblc}[3]{$\vcenter{\hbox{\rotatebox{90}{{\fontfamily{augie}\selectfont\fontsize{7.5}{8}\selectfont \shortstack{#1\\ #2\\ #3}}}}}$}%
\newcommand{\tstrip}[1]{$\vcenter{\hbox{\includegraphics[width=0.46\linewidth]{figures/teaser_frames/#1}}}$}
\newcommand{\dsfont}{\fontfamily{augie}\selectfont\fontsize{7}{8}\selectfont}
\newcommand{\dsimgw}{0.465\textwidth}
\newcommand{\dslbl}[1]{$\vcenter{\hbox{\rotatebox{90}{{\dsfont\shortstack{#1}}}}}$}
\newcommand{\dsimg}[1]{$\vcenter{\hbox{\includegraphics[width=\dsimgw]{figures/dataset/#1}}}$}
\newcommand{\rgfont}{\fontfamily{augie}\selectfont\fontsize{6.5}{7}\selectfont}
\newcommand{\rgcellw}{0.245\textwidth}
\newcommand{\rggap}{3pt}
\definecolor{rgblue}{HTML}{3A6FC4}
\newcommand{\rgcell}[2]{\begin{minipage}[t]{\rgcellw}\centering
  {\rgfont #1}\\[0.5pt]
  {\setlength{\fboxsep}{0pt}\setlength{\fboxrule}{0.6pt}\fcolorbox{rgblue}{white}{\includegraphics[width=\dimexpr\linewidth-1.2pt\relax]{figures/results_grid/#2_in.jpg}}}\\[1pt]
  \includegraphics[width=\linewidth]{figures/results_grid/#2_out.jpg}\end{minipage}}
\newcommand{\rgrowA}[1]{\rgcell{Follow}{#1c0}\hfill\rgcell{Follow Top}{#1c1}\hfill\rgcell{Follow Front}{#1c2}\hfill\rgcell{Follow Behind}{#1c3}}
\newcommand{\rgrowB}[1]{\rgcell{Rotate Back $\rightarrow$ Front}{#1c0}\hfill\rgcell{Rotate Front $\rightarrow$ Back}{#1c1}\hfill\rgcell{Zoom on Face}{#1c2}\hfill\rgcell{Zoom on Figure}{#1c3}}

\twocolumn[{%
\maketitle
\vspace{-4mm}
\begin{center}\begin{minipage}{\textwidth}
\centering
  {\setlength{\tabcolsep}{3pt}\renewcommand{\arraystretch}{1.0}%
  \begin{tabular}{@{}c c @{\hskip 8pt} c c@{}}
  \tlbl{Input} & \tstrip{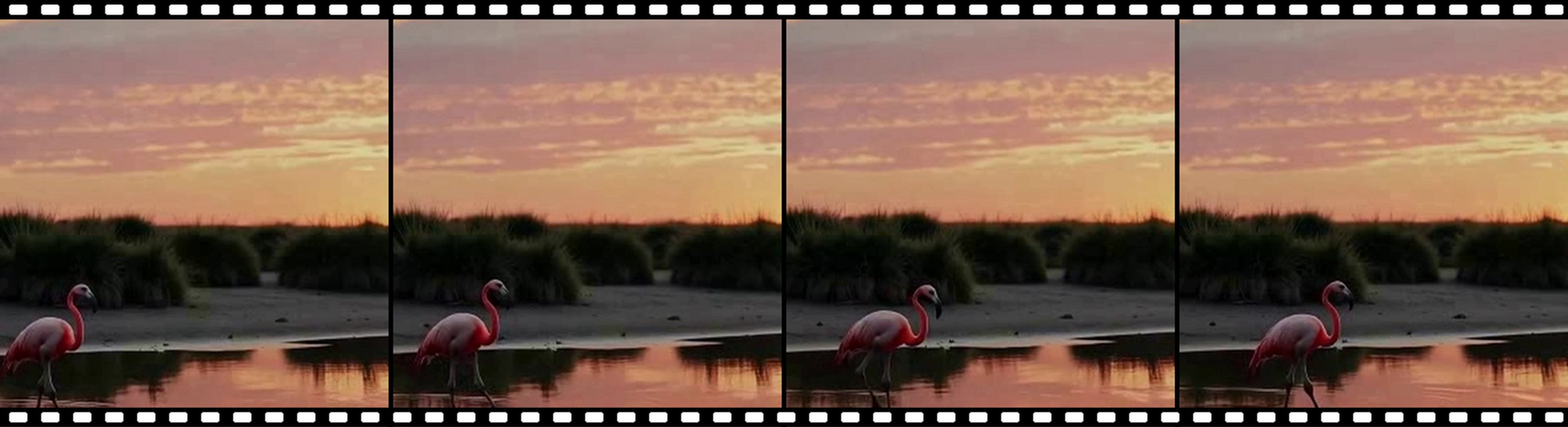} & \tlbl{Input} & \tstrip{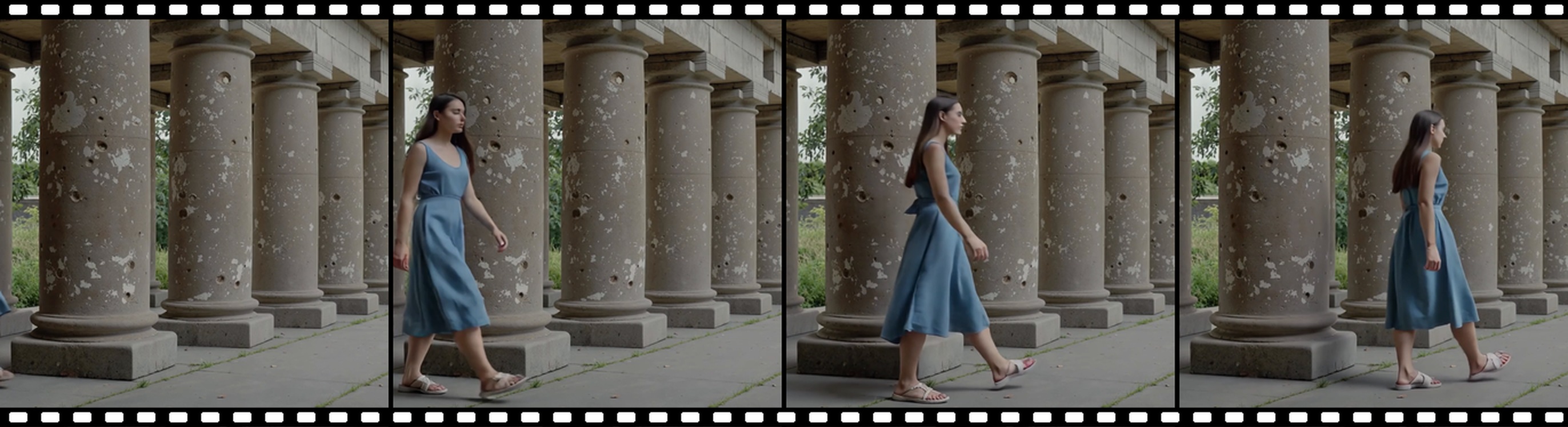} \\[11pt]
  \noalign{\vspace{3pt}}
  \tlbl{Follow} & \tstrip{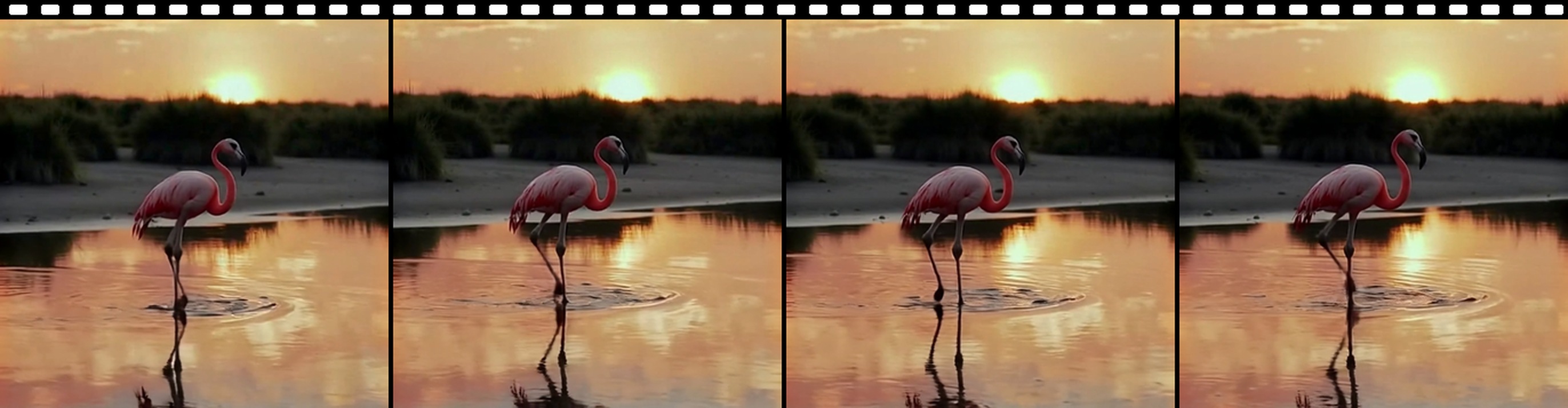} & \tlblb{Rotate}{Back $\rightarrow$ Front} & \tstrip{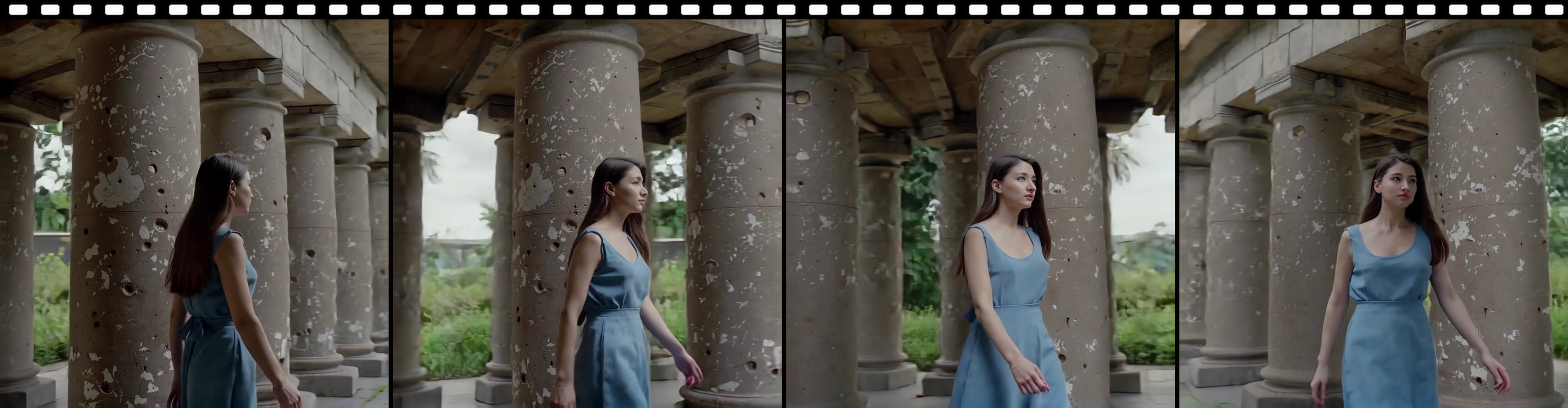} \\[11pt]
  \tlbl{Follow Front} & \tstrip{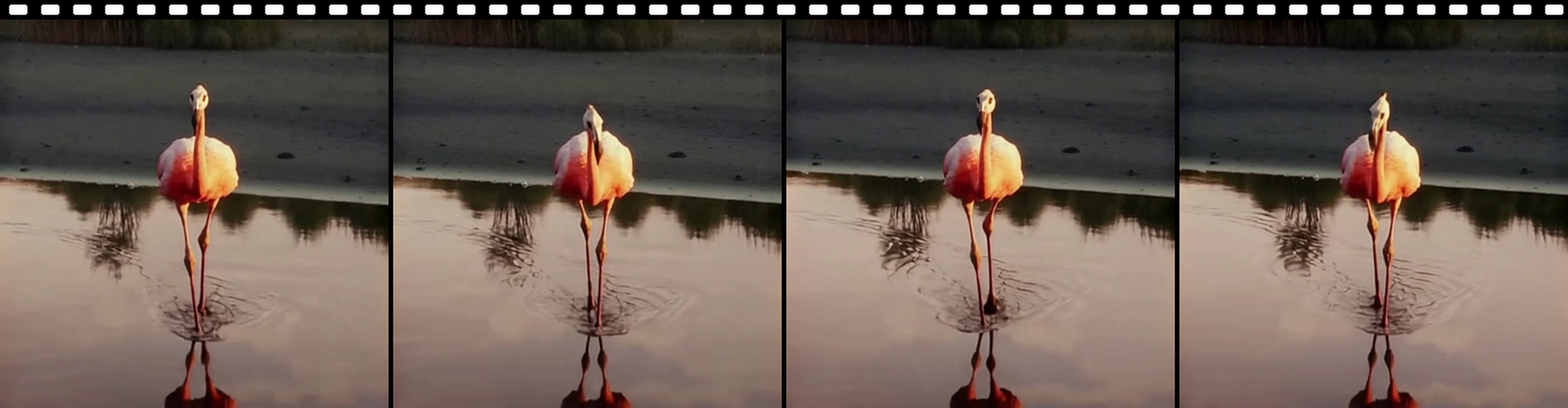} & \tlblb{Rotate}{Front $\rightarrow$ Back} & \tstrip{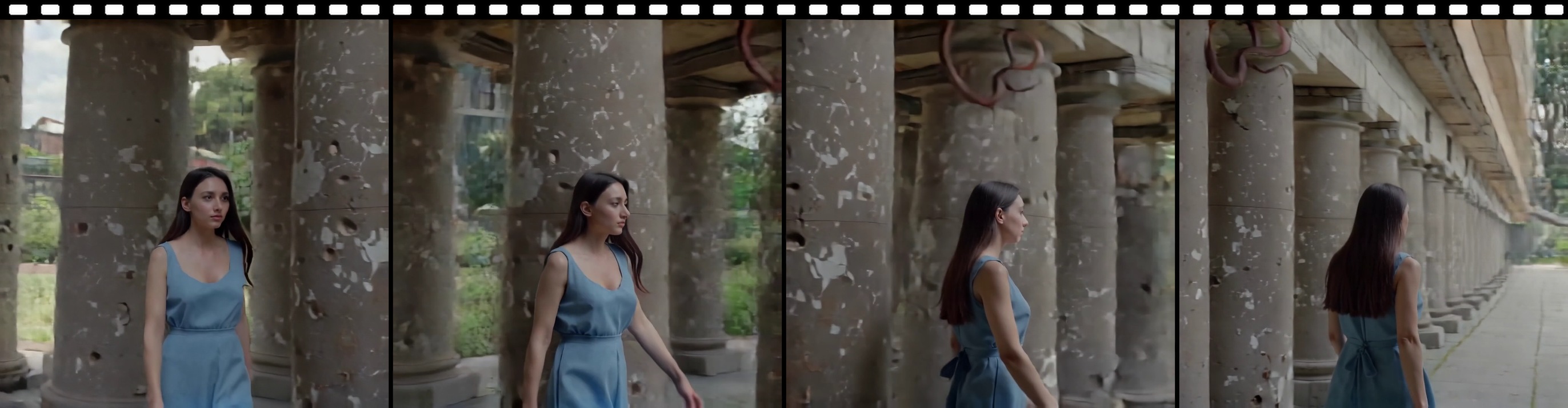} \\[11pt]
  \tlbl{Follow Top} & \tstrip{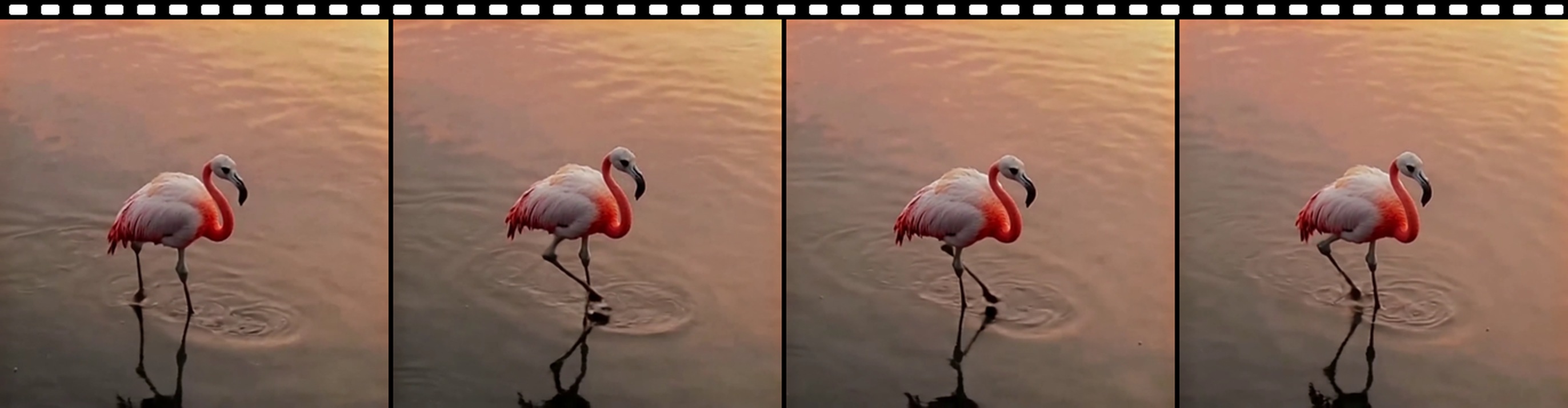} & \tlblc{Rotate $\&$}{Zoom on}{Figure} & \tstrip{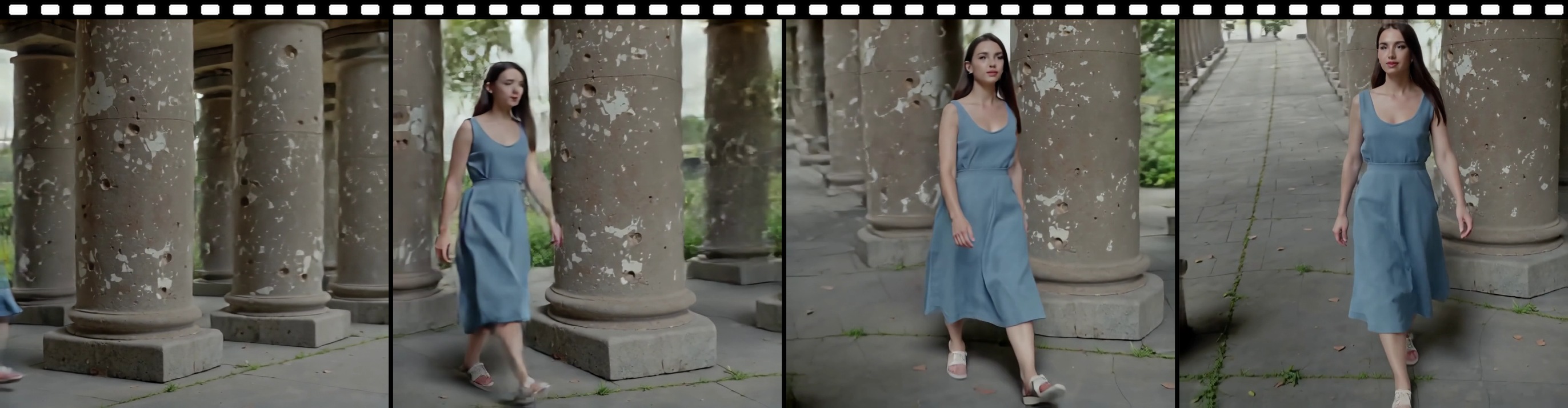} \\[11pt]
  \tlbl{Zoom Face} & \tstrip{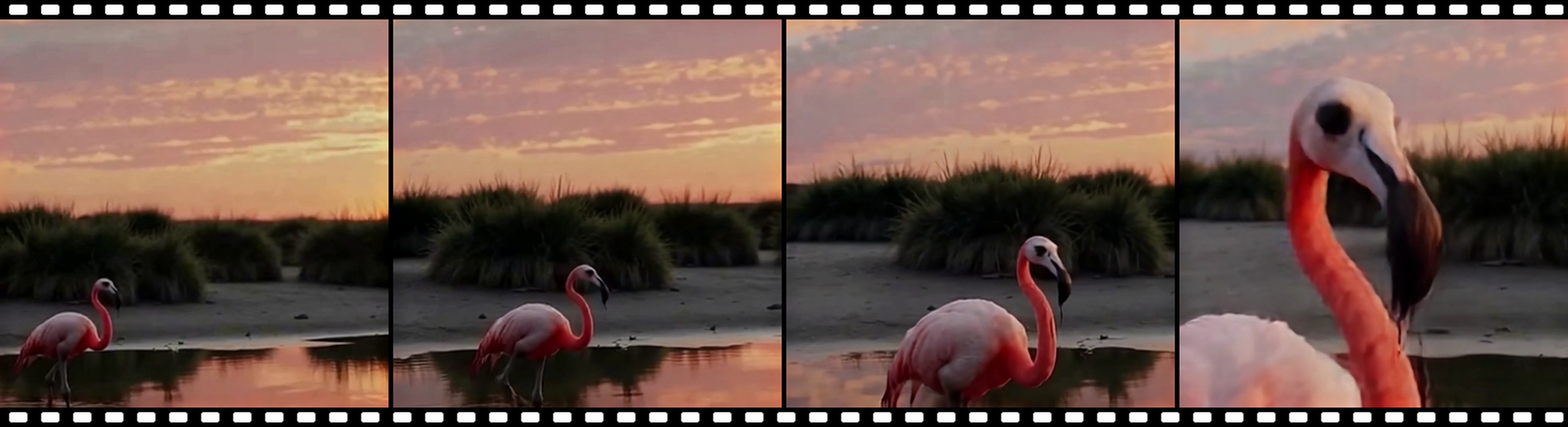} & \tlblb{Follow}{Behind} & \tstrip{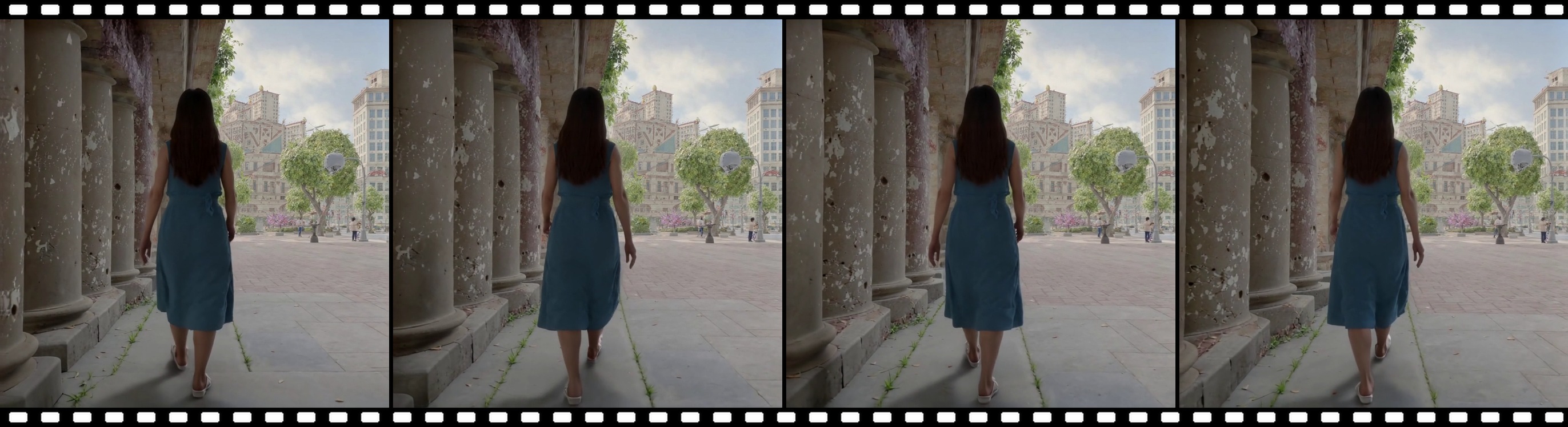} \\
  \end{tabular}}
  \captionof{figure}{\textbf{Semantic camera motion control with \paper.} Given an input video (top row of each block) and a target motion label (left), \paper generates the requested subject-relative camera behavior without a target trajectory, camera poses, or a separate motion-reference video. The camera motion adapts to the scene and the subject's movement.}
  
  \label{fig:teaser}
\end{minipage}\end{center}
\vspace{3mm}}]

\begingroup
\renewcommand{\thefootnote}{\fnsymbol{footnote}}
\footnotetext[1]{Equal contribution.}
\endgroup

\begin{abstract}
Controlling the camera relative to a moving subject in an existing video is challenging: behaviors such as maintaining a frontal view require the camera to adapt to the subject's changing position and orientation, making the desired trajectory difficult to specify in advance. Existing camera-controlled video-to-video methods typically rely on explicit trajectories or reference motions, which do not directly express these dynamic camera--subject relationships. We introduce semantic camera motion control, a novel video-to-video task in which a reference video and a target motion label specify the desired subject-relative camera behavior without an explicit target trajectory. Our method, SemCam, learns to realize this behavior while preserving source content. It combines shared-basis low-rank adaptation with motion-conditioned modulation, while a background-consistency loss encourages fidelity in regions visible in both reference and target videos. We construct 661 paired videos covering eight semantic camera behaviors and evaluate on a separate 109-scene benchmark using subject-relative motion metrics, appearance measures, and a user study. SemCam achieves a semantic-motion success rate of 68.6\%, compared with 45.3\% for Vista4D, the strongest evaluated baseline, while maintaining comparable subject identity preservation.

\end{abstract}

\section{Introduction}

Camera control is essential to filmmaking and visual storytelling. Desired camera behaviors are often expressed relative to a subject: follow a moving person, maintain a frontal view, or approach their face for a close-up. These instructions specify a camera--subject relationship, but the trajectory required to achieve it depends on the scene. Maintaining a frontal view, for example, requires the camera to adapt to the subject's position and heading. The same intention therefore calls for different camera paths as the subject moves, turns, or changes direction.

Existing camera-controlled video generation methods typically require explicit camera trajectories or poses~\cite{he2024cameractrl,wang2024motionctrl,Yu_2025_ICCV,bai2025recammaster}, or a reference video exhibiting the desired motion~\cite{luo2025camclonemaster}. Geometry-based reshooting methods such as Vista4D~\cite{vista4d} likewise require a target camera path. These methods provide control when suitable motion inputs are available, but translating a subject-relative intention into such inputs requires determining an appropriate camera path or reference motion for each scene. A path that maintains the desired framing in one video may lose the subject or expose the wrong side in another.

We introduce \emph{semantic camera motion control}, a novel task in video-to-video generation: given a reference video and a discrete label describing a subject-relative camera behavior, generate a video that realizes that behavior while preserving the subject, action, and scene content. The label specifies what the camera should achieve, while the model infers how the camera should move from the reference video. Unlike classical virtual cinematography, which plans camera motion within an available scene representation~\cite{he1996virtual,galvane2015continuity}, this task requires synthesizing the scene under the requested camera behavior.

This formulation presents two coupled challenges. First, the camera must respond to the subject's evolving position and orientation to maintain the requested relationship throughout the video. Second, changing the viewpoint may reveal parts of the subject and background that are absent from the reference. The model must synthesize these regions plausibly while retaining the identity, action, and visible scene content established by the input. Successful control therefore requires both executing the intended camera behavior and maintaining visual consistency under viewpoint changes.

To address these challenges, we propose \paper, a video diffusion method conditioned on the full reference video and a discrete motion label. Learning distinct behaviors from limited paired data requires balancing information sharing and motion specialization. Independent adapters learn only from the examples available for each behavior, whereas a fully shared adapter must represent all behaviors within the same weight update. We therefore use a factorized low-rank adaptation (LoRA)~\cite{hu2021lora} scheme that combines a basis shared across motions with motion-specific projections. This allows examples from different behaviors to contribute to the shared adaptation while retaining a distinct update for each motion. A complementary modulation branch conditions transformer activations on learned motion embeddings to facilitate learning these behaviors. We further introduce a background-consistency loss focused on regions estimated to be visible in both the reference and target videos, encouraging fidelity to shared background content without imposing this auxiliary constraint on newly revealed regions. The shared factorization also supports composing learned motion factors at inference time.

We collect 661 paired training videos covering eight semantic camera behaviors and evaluate on a separate benchmark of 109 diverse scenes. Because a semantic behavior admits multiple valid camera paths, our subject-relative metrics assess motion execution without requiring a ground-truth target trajectory or output video. Quantitative comparisons demonstrate higher average semantic-motion accuracy than the evaluated camera-control baselines, with subject identity preservation comparable to the strongest baseline. A user study further supports the improvements in camera-control alignment and visual quality.

Our contributions are:
\begin{itemize}
    \item We introduce semantic camera motion control for video-to-video generation, specifying subject-relative camera behavior through a discrete label without requiring an explicit target trajectory.
    \item We propose \paper, combining shared-basis LoRA adaptation, motion-conditioned modulation, and a background-consistency loss, with support for adding and composing camera behaviors.
    \item We construct a paired dataset covering eight semantic camera behaviors and a 109-scene benchmark with subject-relative evaluation metrics, and demonstrate improved motion execution through quantitative comparisons and a user study.
\end{itemize}

\begin{figure*}[t]
  \centering
  \includegraphics[width=0.85\textwidth]{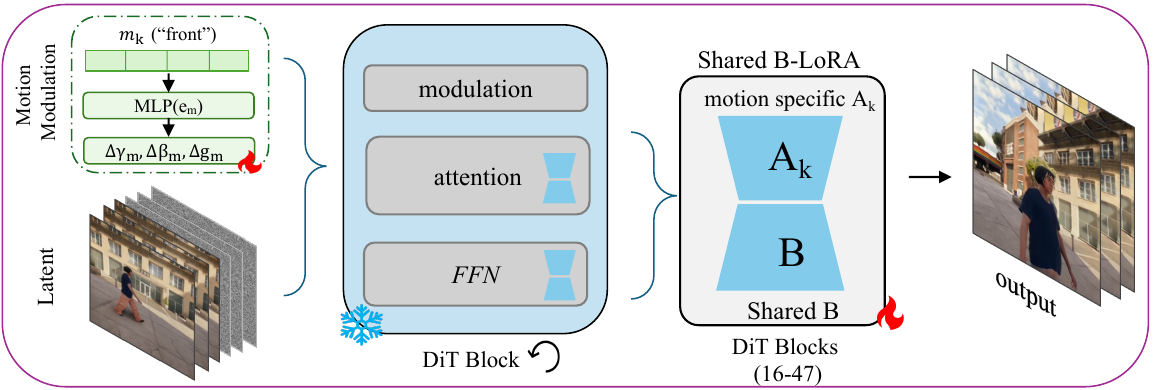}
\caption{\textbf{\paper method overview.} Reference-video tokens are concatenated with noisy target-video tokens $z_t$ and processed by the DiT. The motion label $m$ selects a motion-specific LoRA factor $A_k$, paired with a shared basis $B$, and a learned embedding $e_m$ that predicts modulation residuals $(\Delta\gamma_m,\Delta\beta_m,\Delta g_m)$. These residuals are added to the timestep-dependent modulation parameters. Fire icons indicate trainable components; the pretrained backbone remains frozen.}
  \label{fig:method}
\end{figure*}

\section{Related Work}

\paragraph{Video diffusion models.}
Diffusion models have enabled high-quality video synthesis through both U-Net and transformer architectures, including Video LDM~\cite{blattmann2023videoldm}, Stable Video Diffusion~\cite{blattmann2023stable}, VideoCrafter~\cite{chen2024videocrafter2}, Wan~\cite{wan2025}, and LTX~\cite{hacohen2025ltx2}. These models provide strong generative priors for appearance and temporal dynamics. Our work adapts a pretrained video diffusion model to a specific video-to-video task: changing the camera's behavior relative to a moving subject while retaining the content and action of the reference video. We learn these transformations from paired examples through parameter-efficient adaptation.

\paragraph{Camera-controlled video generation.}
Controllable video generation methods use structural signals such as depth and pose to guide synthesis~\cite{chen2023controlavideo}. Camera-specific approaches condition on camera parameters~\cite{he2024cameractrl,wang2024motionctrl}, combine target trajectories with scene reconstruction or multi-view supervision~\cite{Yu_2025_ICCV,bai2025recammaster,ren2025gen3c,bian2025gsditadvancingvideogeneration}, or transfer camera motion from a reference video~\cite{luo2025camclonemaster}. These approaches offer different interfaces for specifying camera movement, but require a suitable trajectory or motion reference.

Vista4D~\cite{vista4d} reconstructs a video as a 4D point cloud and conditions generation on renderings from target cameras. This provides geometric guidance for preserving scene content, but still requires the desired camera path to be supplied. In our experiments, it also struggles with some viewpoint changes that reveal previously unobserved sides of the subject. We examine these cases alongside other camera-control baselines in our qualitative evaluation.

Related work on autonomous cinematography learns content-dependent camera motion from subject and scene cues, but predicts controls for a physical camera rather than synthesizing video~\cite{huang2019learning}. Our task shares the goal of adapting camera behavior to scene dynamics. However, \paper generates a new video from an input recording and a discrete semantic behavior label, jointly inferring the camera motion and synthesizing the resulting views.

\paragraph{Low-rank adaptation and motion control.}
Low-rank adaptation (LoRA)~\cite{hu2021lora} enables parameter-efficient adaptation of pretrained models. In-Context LoRA~\cite{lhhuang2024iclora} applies task-specific LoRA tuning to exploit the in-context generation capabilities of image diffusion transformers. For video generation, AnimateDiff~\cite{guo2024animatediff} introduces MotionLoRA to learn specific camera motions, while LiON-LoRA~\cite{zhang2025lionlora} studies stable adapter fusion and scalable control of camera and object motion.

Related approaches explore sharing and composing adaptations across transformations. LoRWeB~\cite{manor2026lorweb} learns a basis of LoRA modules and predicts their mixing weights from an input visual analogy, enabling example-based image editing. Our factorization shares parameters within each adapted projection: motion-specific factors $A_m$ are paired with a common factor $B$. This structure allows camera behaviors to share training information while retaining distinct updates, and supports adding new behaviors or mixing learned motion factors.

Modulation offers a complementary mechanism for adaptation. TokenVerse~\cite{garibi2025tokenverseversatilemulticonceptpersonalization} learns token-specific offsets in diffusion-transformer modulation space to extract and compose visual concepts for image personalization. We use learned motion embeddings to predict modulation residuals alongside shared-basis LoRA updates, facilitating the learning of subject-relative camera behaviors from paired videos. We assess the contributions of parameter sharing and modulation through comparisons with independent per-motion adapters, a fully shared adapter, and a variant without motion-conditioned modulation (Sec.~\ref{sec:ablation}).

\section{Method}
\label{sec:method}

Given a reference video $V_{\mathrm{in}}$ and a target camera-motion label $m$, \paper generates a video $V_{\mathrm{out}}$ that realizes the requested subject-relative behavior while preserving the subject, action, and scene content. We learn from paired videos depicting the same scene and action under different camera behaviors, with each pair associated with a motion label.

Our method adapts a pretrained video diffusion transformer through shared-basis LoRA updates and motion-conditioned modulation. The shared basis learns from examples across behaviors, while motion-specific factors retain a distinct adaptation for each behavior. A complementary modulation branch conditions transformer activations on the motion label. We jointly train these components using a flow-matching objective and a background-consistency loss that encourages fidelity in regions estimated to be visible in both reference and target videos. Fig.~\ref{fig:method} illustrates the architecture.

\subsection{Preliminaries}
\label{sec:preliminaries}

\paragraph{Reference-video conditioning.}
We use the reference-video conditioning mechanism implemented in the official LTX-2 IC-LoRA pipeline~\cite{hacohen2025ltx2}. The full reference video is encoded into latent tokens, which are concatenated with noisy target-video tokens along the token dimension and processed jointly by the transformer. This gives the model access to the reference scene and the subject's movement throughout the video. The requested camera behavior is specified separately through our motion-specific adaptation and modulation mechanisms.

The LTX pipeline operates in two stages: low-resolution video generation, followed by latent upsampling and high-resolution refinement. We train our adaptation in the first stage and apply the learned parameters at inference during both stages. Reference-video tokens are concatenated with target tokens in both stages to support content preservation. Implementation details are provided in Appendix~\ref{sec:impl}.
\paragraph{Low-rank adaptation.}
LoRA~\cite{hu2021lora} adapts a pretrained weight matrix $W$ through a low-rank update while keeping $W$ frozen:
\begin{equation}
    W' = W + AB,
\end{equation}
where $A \in \mathbb{R}^{d_{\mathrm{out}}\times r}$, $B \in \mathbb{R}^{r\times d_{\mathrm{in}}}$, and $r \ll \min(d_{\mathrm{in}},d_{\mathrm{out}})$. This reduces the number of trainable parameters compared with full fine-tuning. We apply these updates to attention projections in selected transformer blocks, omitting layer indices below for clarity.

\paragraph{Feature-wise modulation.}
Feature-wise modulation conditions network activations through learned scale and shift parameters~\cite{perez2018film}. Diffusion transformers use related mechanisms to condition their blocks on the denoising timestep~\cite{10377858}. In our LTX backbone~\cite{hacohen2025ltx2}, the attention and feed-forward branches use timestep-dependent scale, shift, and residual-gating parameters, denoted by $\gamma_b^t$, $\beta_b^t$, and $g_b^t$, respectively, where $b\in\{\mathrm{attn},\mathrm{ffn}\}$.

\subsection{\paper}
\label{sec:shared_lora}

\paragraph{Shared-basis motion LoRA.}
Our adaptation separates parameters shared across camera behaviors from those specific to each behavior. For each adapted projection, we learn a common basis matrix $B$ and a motion-specific factor $A_m$, selected by the requested label $m$:
\begin{equation}
    W'_m = W + A_mB.
\end{equation}
During joint training, examples from all motion categories update $B$, while each example updates only its corresponding factor $A_m$. This lets different behaviors contribute to the shared representation while retaining a distinct weight update for each motion.

This factorization provides a middle ground between independent adapters, $W'_m=W+A_mB_m$, and a fully shared adapter, $W'=W+AB$. Independent adapters learn only from the examples available for their respective behaviors, whereas a fully shared adapter uses the same weight update for every behavior. In our experiments with limited paired data, independent adapters generalize less effectively to unseen scenes, while the fully shared adapter often produces weak or incorrect camera motion (Sec.~\ref{sec:ablation}). Sharing $B$ while retaining motion-specific factors addresses this trade-off between sharing information and specializing the adaptation.

The common basis $B$ also supports combining motion factors at inference time and learning additional behaviors while keeping the shared basis frozen. We evaluate these capabilities in Sec.~\ref{sec:fusion}.

\begin{table*}[t]
\centering
\caption{\textbf{Comparison with camera-control baselines} on 109 scenes and eight semantic camera target motions (872 clips per method). Scores are averaged within each motion type and then across motions. \emph{Sem.\ motion} denotes the execution score $M$; \emph{Success} is the fraction of clips with $M>0.5$. DINO and ID-Sim measure subject appearance relative to the input, while the VLM judge evaluates motion execution and identity preservation accounting for viewpoint changes.}
\label{tab:consolidated_updated}
\begin{tabular}{lcccccc}
\toprule
& \multicolumn{2}{c}{Motion} & \multicolumn{2}{c}{Subject appearance} & \multicolumn{2}{c}{VLM judge} \\
\cmidrule(lr){2-3}\cmidrule(lr){4-5}\cmidrule(lr){6-7}
Method & Sem.\ motion~$\uparrow$ & Success~$\uparrow$ & DINO~$\uparrow$ & ID-Sim~$\downarrow$ & Sem.\ motion~$\uparrow$ & Identity~$\uparrow$ \\
\midrule
TrajectoryCrafter   & 0.238 & 0.223 & 0.426 & 0.506 & 0.377 & 0.520 \\
ReCamMaster         & 0.139 & 0.138 & 0.531 & 0.407 & 0.281 & 0.725 \\
CamCloneMaster      & 0.180 & 0.172 & 0.591 & 0.329 & 0.299 & 0.732 \\
Vista4D             & 0.429 & 0.453 & 0.592 & 0.307 & 0.476 & \textbf{0.750} \\
\midrule
\textbf{Ours}       & \textbf{0.642} & \textbf{0.686} & \textbf{0.594} & \textbf{0.293} & \textbf{0.621} & \textbf{0.750} \\
\bottomrule
\end{tabular}
\end{table*}

\begin{figure*}[p]
    \centering
    \includegraphics[width=0.4\textwidth]{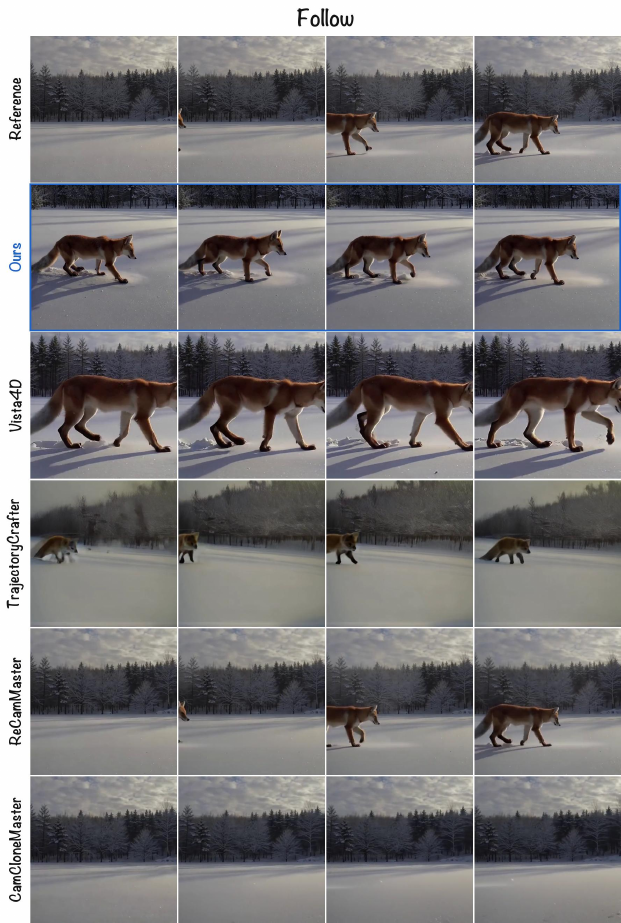}
    \includegraphics[width=0.4\textwidth]{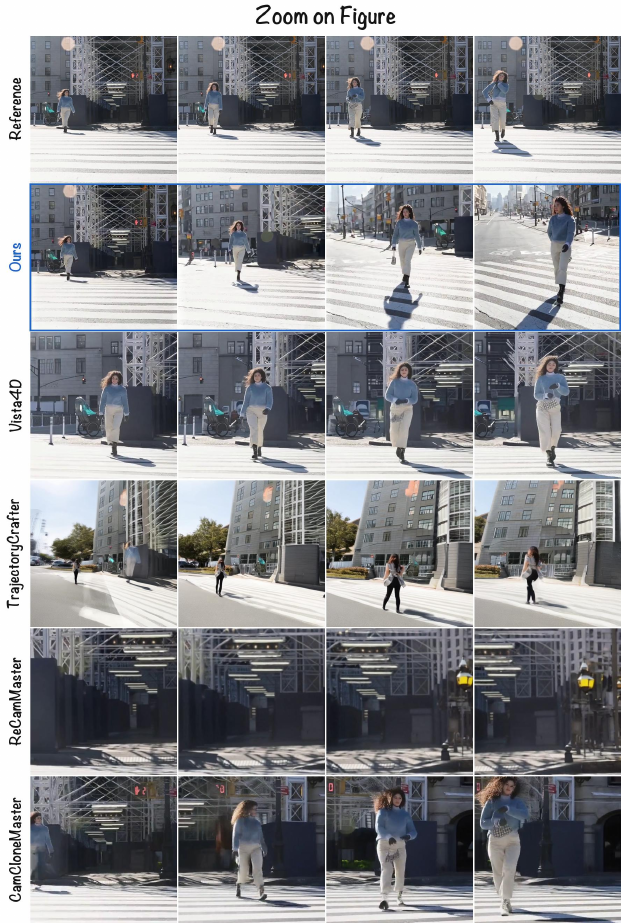}\\[4pt]
    \includegraphics[width=0.4\textwidth]{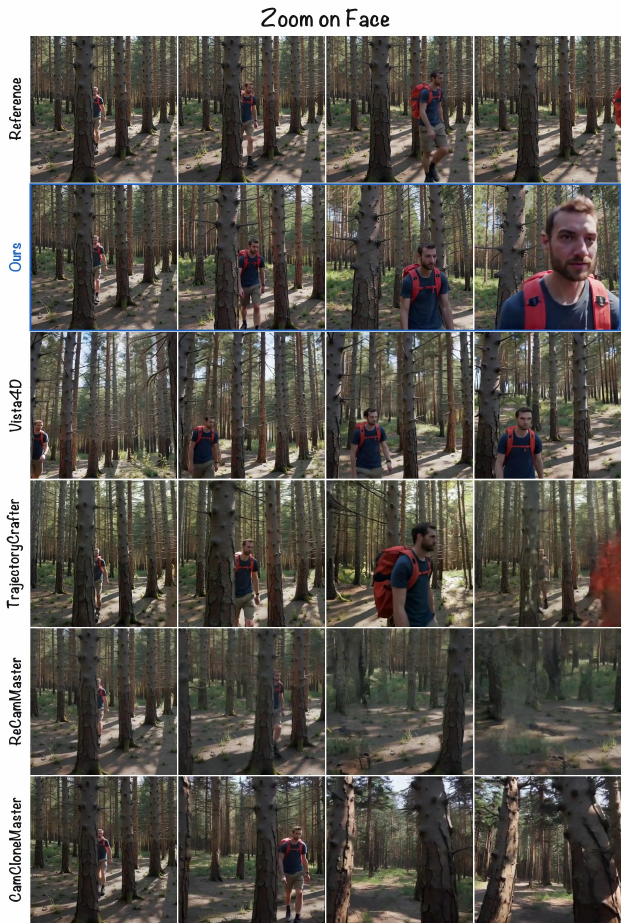}
    \includegraphics[width=0.4\textwidth]{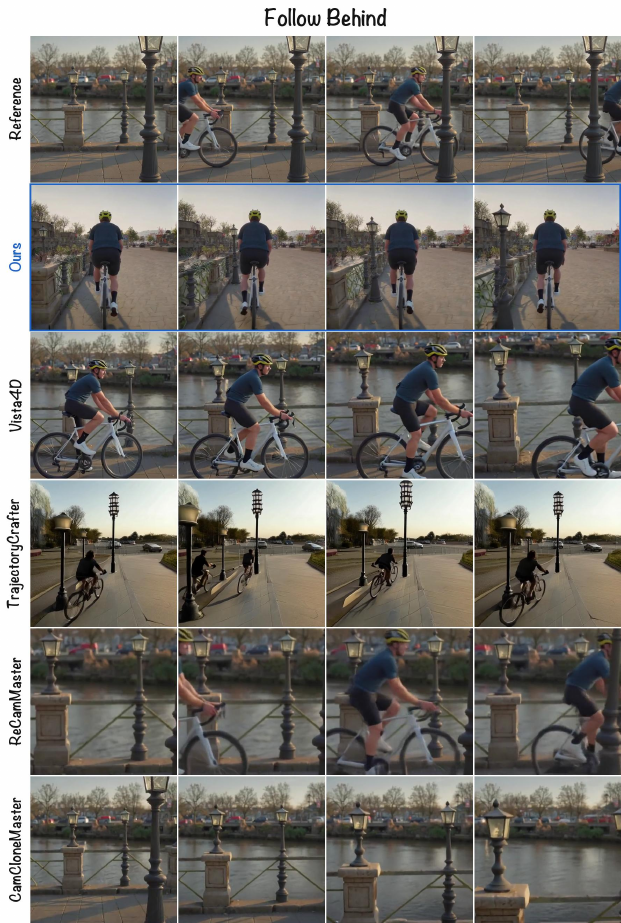}
\caption{\textbf{Qualitative comparison with camera-control baselines.} Four examples compare \paper (highlighted in blue) with Vista4D~\cite{vista4d}, TrajectoryCrafter~\cite{Yu_2025_ICCV}, ReCamMaster~\cite{bai2025recammaster}, and CamCloneMaster~\cite{luo2025camclonemaster}. Each example is labeled by the requested camera behavior; rows show the input and method outputs, and columns show four frames sampled over time.}
    \label{fig:baseline_full}
\end{figure*}

\paragraph{Motion-conditioned modulation.}
\label{sec:motion_modulation}
To facilitate learning distinct camera behaviors, we introduce a motion-conditioned modulation branch. Each motion label $m$ is associated with a trainable embedding $e_m$, which a lightweight MLP maps to residual scale, shift, and gating parameters for the attention and feed-forward branches:

\begin{equation}
\begin{aligned}
\mathrm{MLP}(e_m) = (&\Delta\gamma_{\mathrm{attn},m},
\Delta\beta_{\mathrm{attn},m},
\Delta g_{\mathrm{attn},m},\\
&\Delta\gamma_{\mathrm{ffn},m},
\Delta\beta_{\mathrm{ffn},m},
\Delta g_{\mathrm{ffn},m}).
\end{aligned}
\end{equation}
We add these residuals to the backbone's timestep-dependent parameters:
\begin{equation}
\begin{aligned}
    \gamma_b &= \gamma_b^t + \Delta\gamma_{b,m},\\
    \beta_b &= \beta_b^t + \Delta\beta_{b,m},\\
    g_b &= g_b^t + \Delta g_{b,m},
\end{aligned}
\qquad b\in\{\mathrm{attn},\mathrm{ffn}\}.
\end{equation}

The two adaptation pathways act at different points in the transformer: $A_mB$ modifies attention projection weights, while the modulation branch adjusts feature activations and residual contributions. Jointly training both pathways improves motion execution compared with shared-basis LoRA alone (Sec.~\ref{sec:ablation}).

\paragraph{Flow-matching objective.}
\label{sec:training_objective}
Each training example consists of a reference video $V_{\mathrm{in}}$, a target video $V^\star$, and a motion label $m$. Let $z_t$ denote the noisy target latent at timestep $t$. The model predicts the flow-matching velocity conditioned on the reference video and motion label:
\begin{equation}
    \mathcal{L}_{\mathrm{flow}}
    =
    \mathbb{E}
    \left[
    \left\|
    v_\theta(z_t,t,V_{\mathrm{in}},m)-v^\star
    \right\|_2^2
    \right],
\end{equation}
where $v^\star$ is the target velocity defined by the backbone's flow-matching formulation. The expectation is over training examples, timesteps, and sampled noise. We optimize the motion-specific factors $\{A_m\}$, shared bases $B$, motion embeddings, and modulation MLP while keeping the pretrained backbone frozen.

\paragraph{Background-consistency loss.}
\label{sec:bg_loss}
The flow-matching objective supervises velocity prediction across the full latent video, but does not explicitly prioritize semantic agreement in background regions shared with the reference. We therefore add a feature-space loss focused on these regions.

A frozen convolutional feature predictor $h$, trained offline to approximate DINOv2~\cite{oquab2023dinov2} patch features from video latents, maps the denoised target estimate $\hat z_0$ into semantic features. We compare its predictions with precomputed DINOv2 features $\phi(V^\star)$ of the ground-truth target video. This allows us to compute the feature-space loss without decoding the predicted video or running DINOv2 on it during training.

We restrict the comparison to a mask $\mathcal{M}$ of target patches estimated to depict background visible in both videos. The mask is derived from joint reconstruction of the reference and target using STream3R~\cite{stream3r2025}, followed by reprojection-based visibility estimation and exclusion of dynamic-subject regions. The loss is
\begin{equation}
    \mathcal{L}_{\mathrm{bg}}
    =
    \frac{1}{|\mathcal{M}|}
    \sum_{p\in\mathcal{M}}
    \left[
    1-\cos\left(
    h(\hat z_0)_p,\phi(V^\star)_p
    \right)
    \right],
\end{equation}
where $p$ indexes patches across video frames. We set this loss to zero when $\mathcal{M}$ is empty.

The mask focuses the additional supervision on shared background content. Newly revealed regions remain supervised by the flow-matching objective but are excluded from this auxiliary constraint. Unlike alignment of internal model representations~\cite{yu2025repa}, our loss supervises features predicted from the denoised output estimate; unlike whole-frame latent perceptual objectives~\cite{berrada2025lpl}, it targets estimated shared background regions.

The total training objective is
\begin{equation}
    \mathcal{L}_{\mathrm{total}}
    =
    \mathcal{L}_{\mathrm{flow}}
    +\lambda w(t)\mathcal{L}_{\mathrm{bg}},
\end{equation}
where $\lambda=0.5$ and $w(t)$ suppresses the auxiliary loss at high noise levels, when the denoised estimate is unreliable. The weighting schedule and implementation details are provided in Appendix~\ref{sec:impl}; the visibility mask is analyzed in Appendix~\ref{sec:bg_loss_appendix}.

\begin{figure*}[th!]
    \centering
    \rgrowA{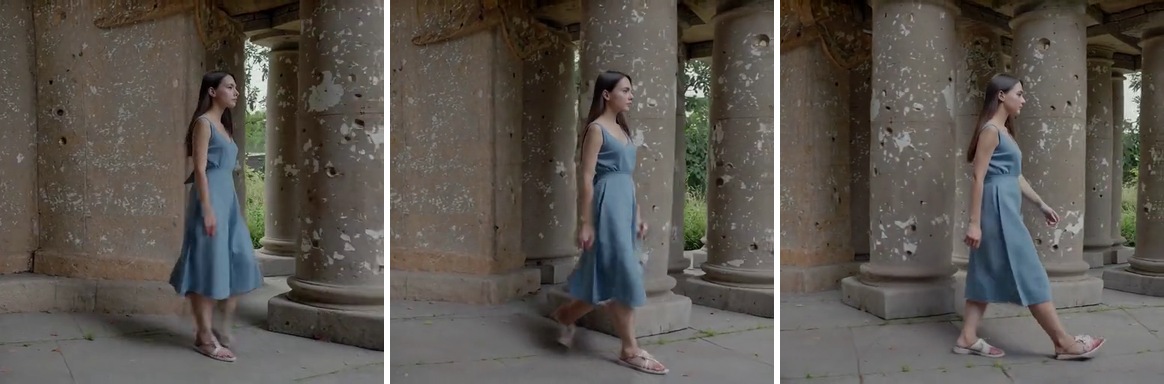}\\[\rggap]
    \rgrowB{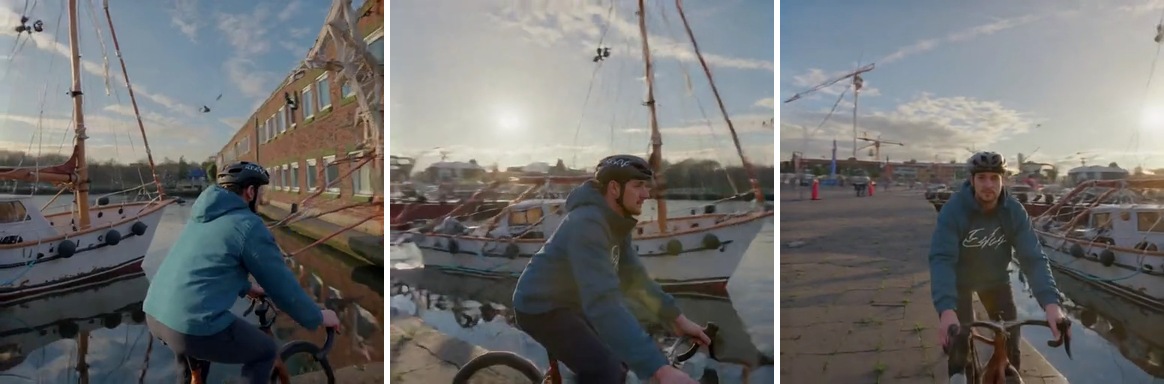}\\[\rggap]
    \rgrowA{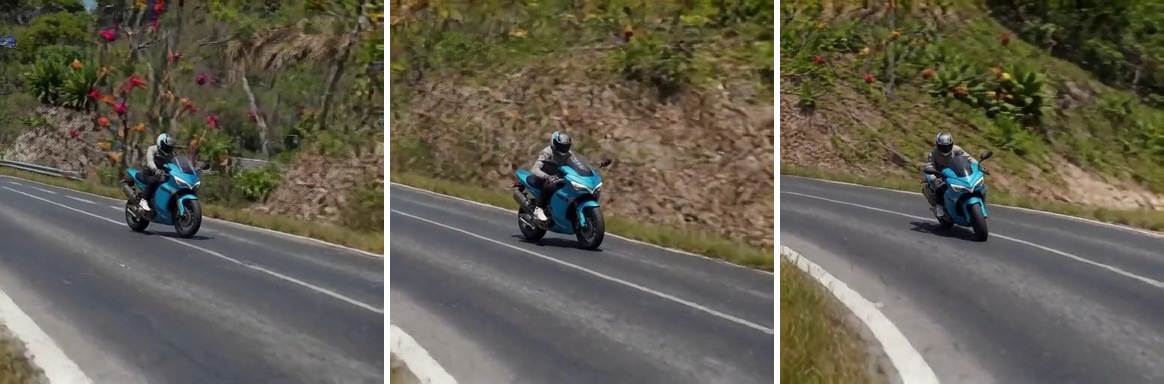}\\[\rggap]
    \rgrowB{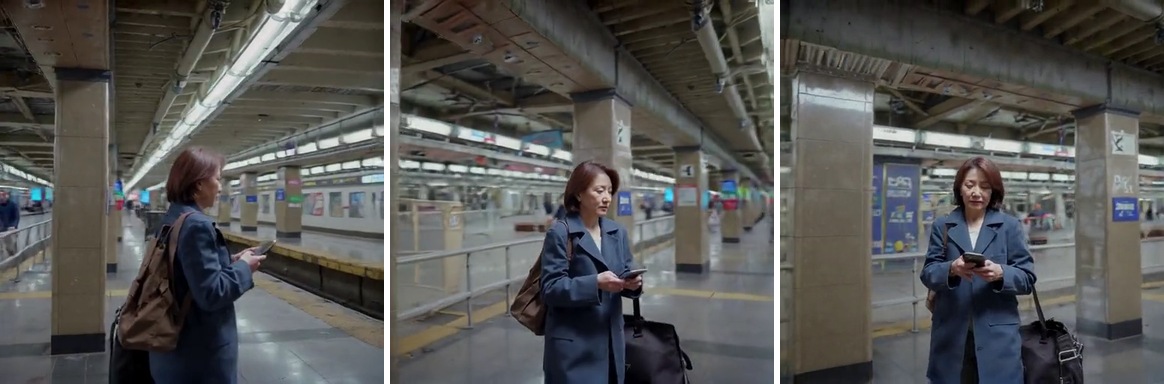}
\caption{\textbf{Semantic camera control across diverse scenes.} Each cell shows three frames from the input video (top, blue) and \paper's output (bottom) for the labeled camera behavior. Outputs are generated independently from each input video and motion label. Examples span people, animals, vehicles, and stylized subjects; all clips are included in the supplementary videos.}
    \label{fig:results}
\end{figure*}

\section{Experiments}
We evaluate whether \paper executes the requested subject-relative camera behavior while preserving the appearance of the input subject. We compare against camera-controlled video generation methods, assess agreement with human preferences, and ablate the main components of our adaptation. We additionally investigate whether learned motion factors can be composed at inference time.
\begin{figure}[t]
  \centering
  \includegraphics[width=\linewidth]{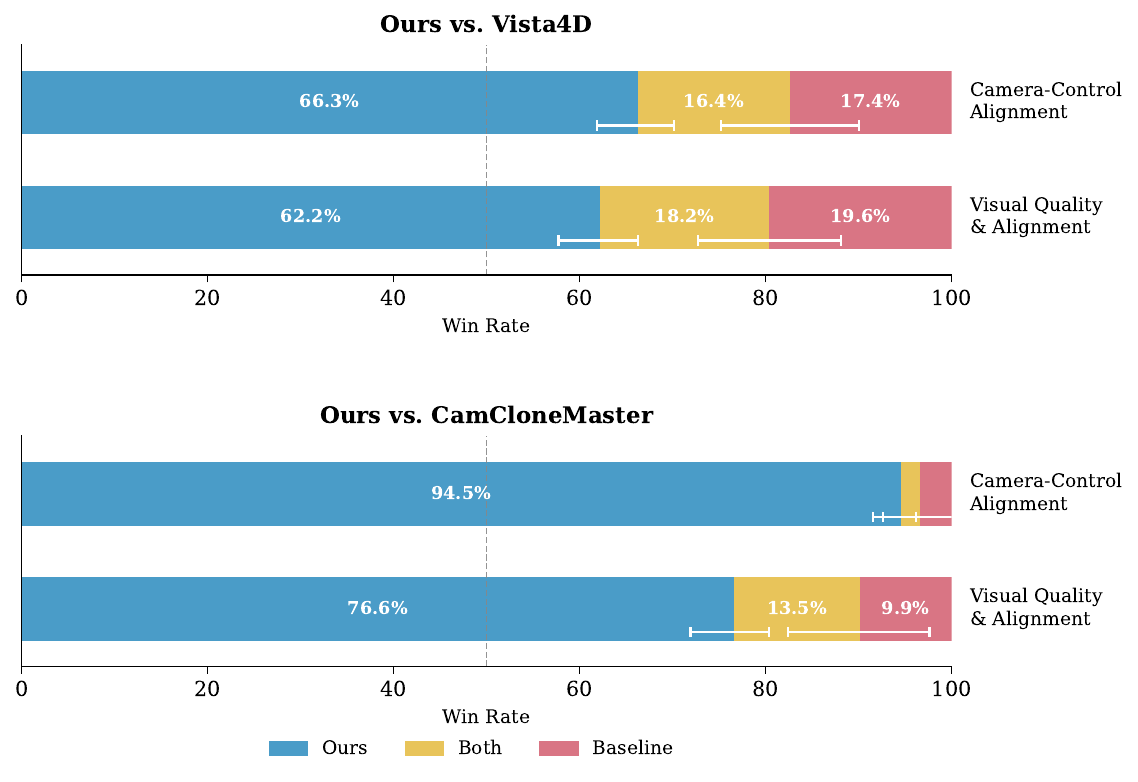}
  \caption{\textbf{User study} comparing our method against  Vista4D~\cite{vista4d} (top) and CamCloneMaster~\cite{luo2025camclonemaster} (bottom), conducted with 55 participants. Each horizontal bar shows the fraction of responses preferring our method (\textcolor[HTML]{4A9CC8}{\rule{0.8em}{0.8em}}~Ours), rating both equally (\textcolor[HTML]{E8C45A}{\rule{0.8em}{0.8em}}~Both), or preferring the baseline (\textcolor[HTML]{D97584}{\rule{0.8em}{0.8em}}~Baseline) for camera-control alignment (top) and visual quality (bottom).}
  \label{fig:user_study}
\end{figure}

\subsection{Experimental Setup}

\paragraph{Training dataset.}
\label{sec:dataset}
We construct a dataset of 661 paired videos covering eight semantic camera behaviors: \emph{follow}, \emph{follow front view}, \emph{follow top view}, \emph{follow from behind}, \emph{rotate back-to-front}, \emph{rotate front-to-back}, \emph{rotate and zoom on face}, and \emph{rotate and zoom on figure}. Each pair depicts the same scene and subject action under different camera behaviors. Example input--target video pairs are shown in Fig.~\ref{fig:dataset_samples}.

We collect footage from Skate~\cite{skate2025} and GTA~V~\cite{gta5}, which allow the same action to be captured from different camera viewpoints. The dataset primarily features human subjects, with a small number of animal examples. Per-motion statistics are provided in Tab.~\ref{tab:dataset_v9}, and representative video pairs are shown in Fig.~\ref{fig:dataset_samples}.

\paragraph{Evaluation benchmark.}
\label{sec:eval_benchmark}
Our evaluation benchmark contains 109 scenes drawn from real-world footage and LTX-generated videos. It includes humans, animals, vehicles, robots, and stylized subjects, with approximately half of the scenes containing subject categories absent from training. The benchmark also includes challenging cases involving multiple subjects, occlusion, abrupt direction changes, fast movement, and subjects entering or leaving the frame. We evaluate all eight behaviors for each scene, yielding 872 output videos per method.

\paragraph{Baselines.}
\label{sec:baselines}
We compare against Vista4D~\cite{vista4d}, TrajectoryCrafter~\cite{Yu_2025_ICCV}, ReCamMaster~\cite{bai2025recammaster}, and CamCloneMaster~\cite{luo2025camclonemaster}. Vista4D and TrajectoryCrafter use a supplied camera trajectory to render geometric guidance for video generation, while ReCamMaster directly conditions on target camera poses. CamCloneMaster instead receives a separate video demonstrating the desired camera motion.

To translate semantic behaviors into these inputs, we author one canonical trajectory per behavior and convert it to each trajectory-conditioned method's camera convention. For CamCloneMaster, we provide a camera-motion reference depicting the requested behavior. \paper receives the input video and a discrete label specifying the camera behavior. We evaluate all methods on the same scene--behavior pairs, measuring motion execution and subject appearance preservation.

We evaluate complete output clips at each method's native resolution and aspect ratio; output formats and evaluation sampling are detailed in Appendix~\ref{sec:baseline_protocol}.

\subsection{Evaluation Metrics}
\label{sec:metrics}

\begin{figure}[!t]
    \centering
    \definecolor{ablref}{RGB}{95,175,70}
    \newcommand{\ablf}{\fontfamily{augie}\selectfont}
    \newcommand{\ablrow}[1]{\rotatebox{90}{\makebox[0.4988\ablimgw][c]{\ablf\fontsize{6}{5.5}\selectfont #1}}}
    \newcommand{\ablimgw}{\dimexpr0.5\linewidth-4.5pt\relax}%
    \newcommand{\ablimg}[1]{\includegraphics[width=\ablimgw]{figures/ablation_frames/#1.jpg}}
    \setlength{\tabcolsep}{0pt}\renewcommand{\arraystretch}{0.4}\setlength{\fboxsep}{0pt}\setlength{\fboxrule}{0.8pt}
    \begin{tabular}{@{}c@{\hspace{1.5pt}}c@{\hspace{1pt}}c@{}}
     & {\ablf\fontsize{8.5}{8.5}\selectfont Follow Top} & {\ablf\fontsize{8.5}{8.5}\selectfont Follow Front}\\[1.5pt]
    \ablrow{Input} & \ablimg{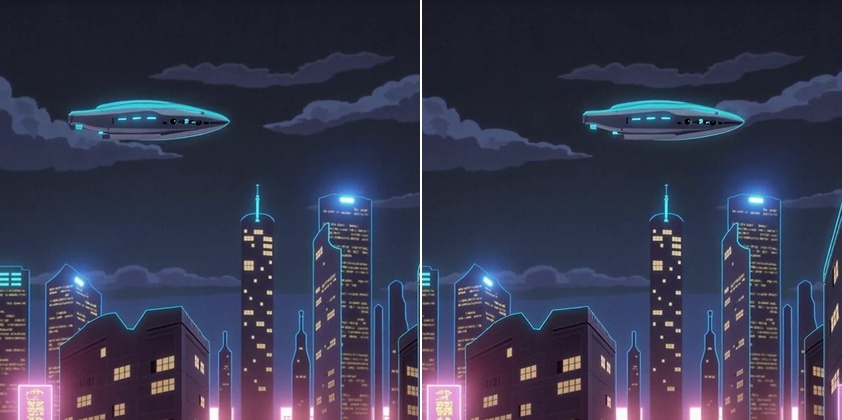} & \ablimg{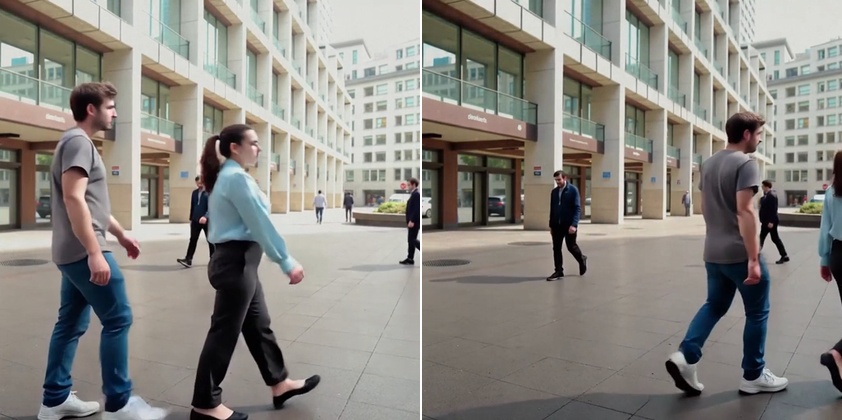}\\[1pt]
    \ablrow{Full method} & \ablimg{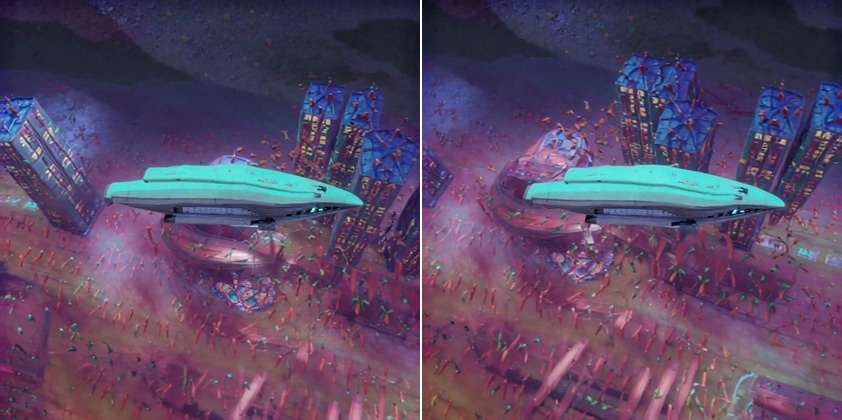} & \ablimg{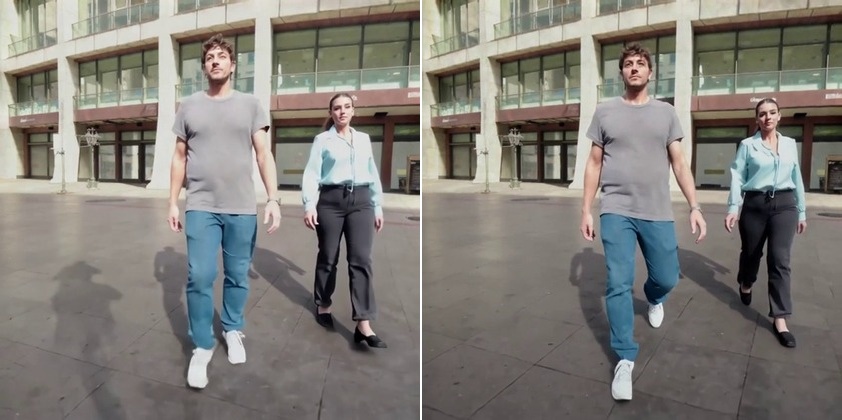}\\[1pt]
    \ablrow{-- background loss} & \ablimg{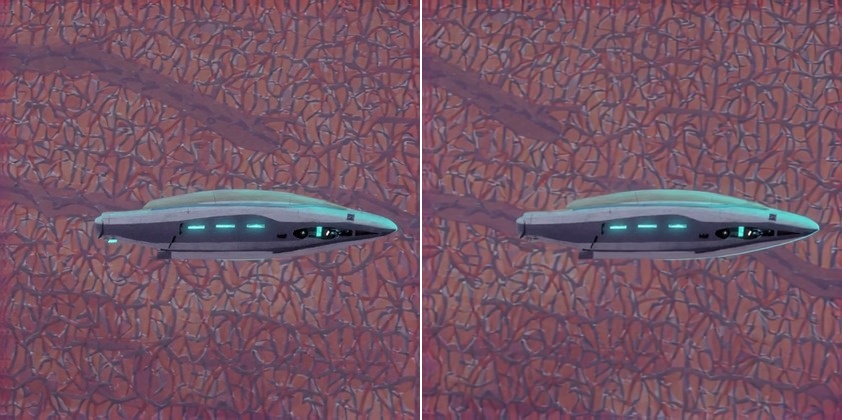} & \ablimg{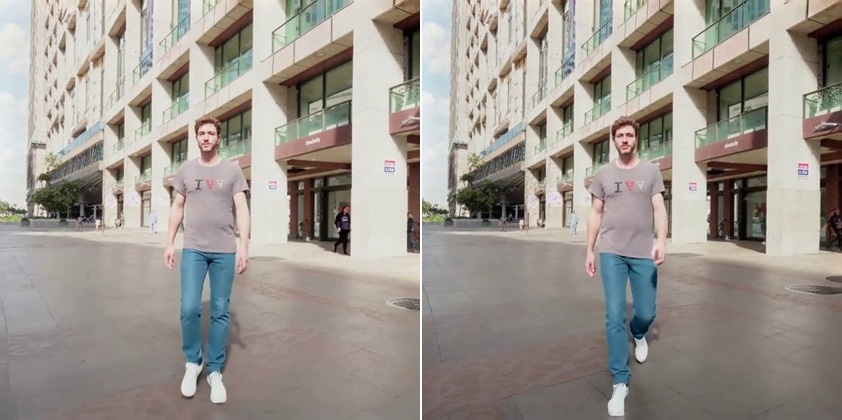}\\[1pt]
    \ablrow{-- modulation} & \ablimg{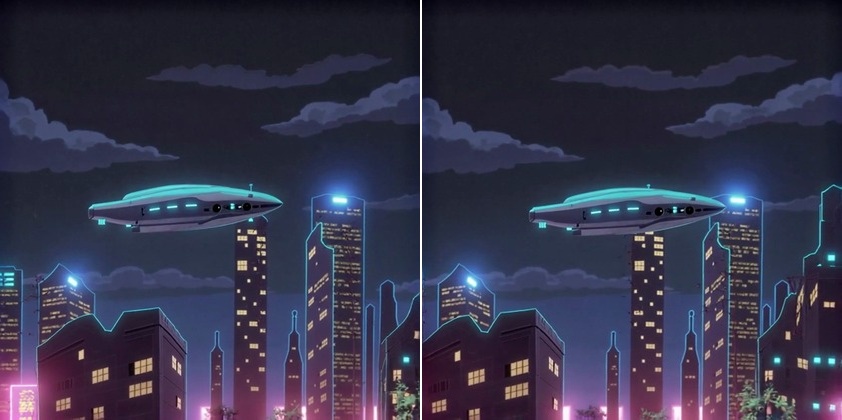} & \ablimg{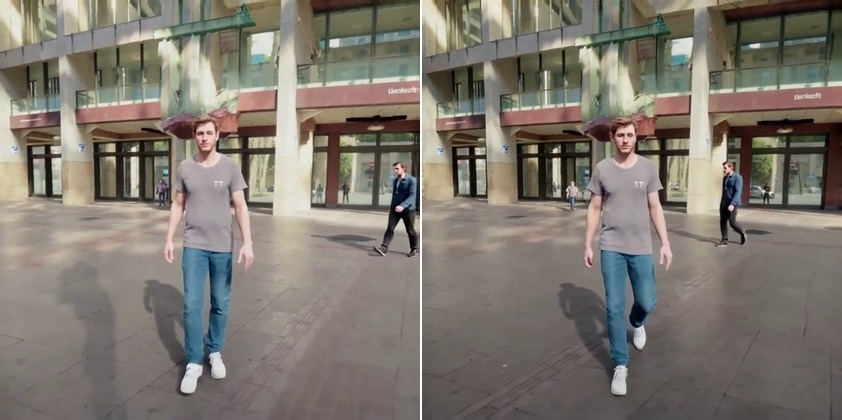}\\[1pt]
    \ablrow{Shared LoRA $AB$} & \ablimg{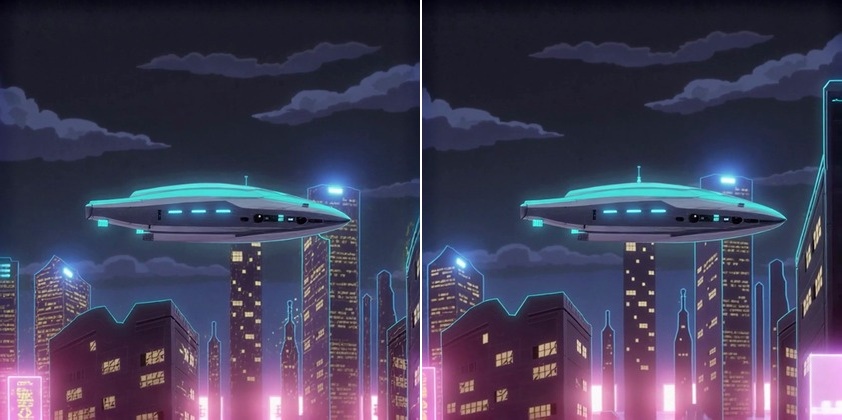} & \ablimg{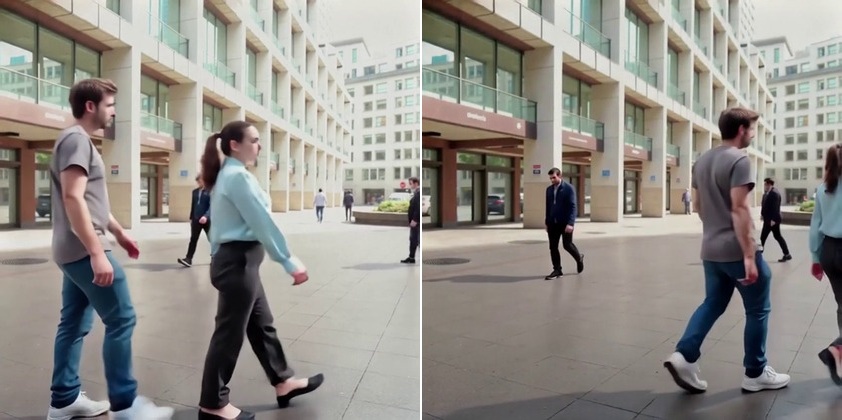}\\[1pt]
    \ablrow{Per-motion LoRA} & \ablimg{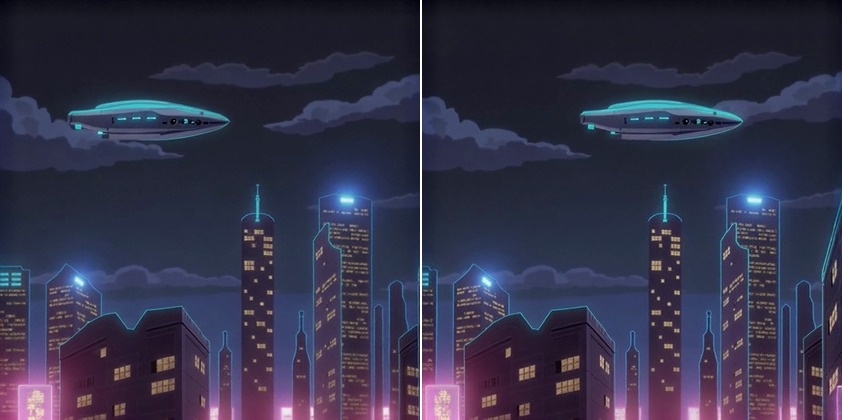} & \ablimg{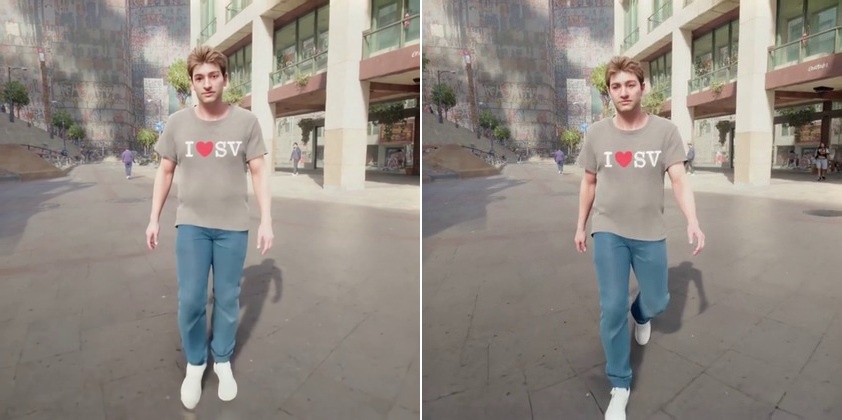}\\[1pt]
    \ablrow{Text-conditioned} & \ablimg{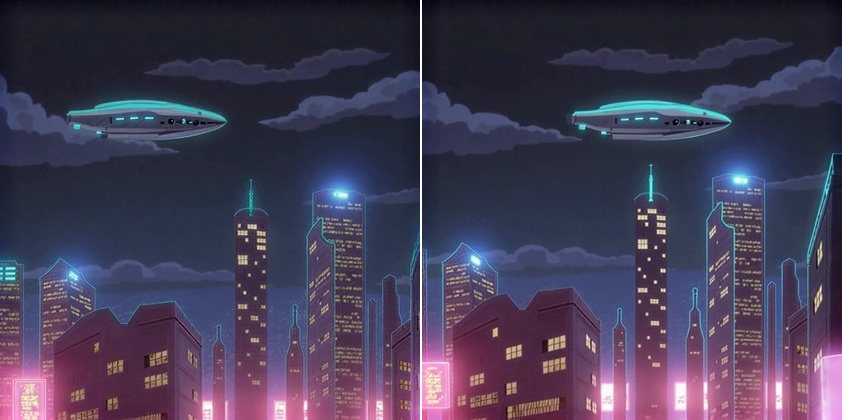} & \ablimg{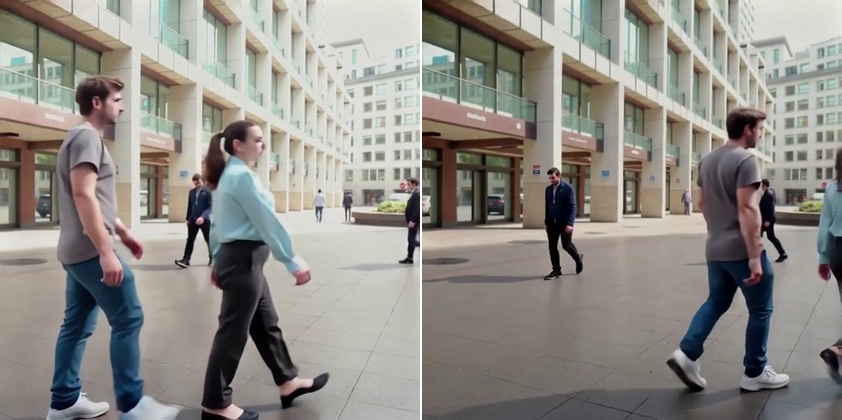}\\
    \end{tabular}
    \caption{\textbf{Ablation study.} Two semantic camera motions  (columns; two frames per clip) for the full model and each ablation of Tab.~\ref{tab:ablation_main}. Our full model both executes the requested camera motion control and preserves the scene.}
    \label{fig:ablation}
\end{figure}

A semantic camera instruction admits multiple valid outputs, making pixel-wise comparison with a single target video inappropriate. We therefore assess the requested camera--subject relationship directly and evaluate subject appearance separately. We complement these measurements with a VLM judge and a human user study.
\paragraph{Semantic-motion execution.}
We sample 25 frames from each output and estimate camera poses, intrinsics, and depth using Depth Anything~3~\cite{lin2025depthanything3}. We segment the subject with SAM~3~\cite{carion2025sam3} and estimate its 3D position by back-projecting its mask centroid at the median subject depth. These estimates provide camera--subject distance, relative camera movement, and subject magnification. Orient Anything~V2~\cite{wangorient} estimates subject orientation and viewing elevation from subject crops.

We derive seven components, each normalized to $[0,1]$, with higher values indicating better satisfaction of the corresponding criterion: \emph{travel}, \emph{distance kept}, \emph{elevation}, \emph{view}, \emph{zoom}, \emph{face at end}, and \emph{figure framing}. Each behavior requires a subset of these components (Tab.~\ref{tab:motion_components}). We define the semantic-motion score as
\begin{equation}
    M = \min_{c\in\mathcal{C}(m)} c,
\end{equation}
where $\mathcal{C}(m)$ contains the components required by behavior $m$. Taking the minimum prevents strong performance on one component from compensating for failure on another.

For example, \emph{follow front} combines \emph{travel}, \emph{distance kept}, and \emph{view}: the camera must move, maintain approximately constant distance from the subject, and retain a frontal view throughout the clip. Illustrative component scores of $0.8$, $0.7$, and $0.3$ yield $M=0.3$, indicating that the frontal-view criterion limits overall execution despite stronger movement and distance scores. Conversely, a static frontal view is penalized through the travel component.

We count a clip as successful when $M>0.5$, requiring every component associated with the requested behavior to exceed $0.5$. Among clips successfully processed by the evaluator, those with the subject detected in fewer than 30\% of sampled frames or absent at either endpoint receive zero and remain in the averages. Scores and success rates are averaged within each behavior and then across behaviors. We report both the continuous score $M$, which captures partial satisfaction, and the success rate, which measures how often all required criteria exceed the threshold.

\paragraph{Subject appearance.}
We measure subject preservation using DINOv2~\cite{oquab2023dinov2} cosine similarity and ID-Sim~\cite{Chae_2026_CVPR}, a viewpoint-robust identity distance, computed on subject crops extracted with GroundingDINO~\cite{liu2024groundingdino}. Higher DINO similarity and lower ID-Sim distance indicate greater appearance agreement with the reference. We interpret these measures alongside motion execution, since outputs with little camera movement can retain high reference similarity.

\begin{table*}[t]
\centering
\caption{\textbf{Success rate per semantic motion} ($M>0.5$), 109 scenes per motion. Best per column in \textbf{bold}.}
\label{tab:motion_specific}
\small
\begin{tabular}{lcccccccc}
\toprule
Method & Follow & Top & Front & Behind & Back$\to$front & Front$\to$back & Zoom face & Zoom figure \\
\midrule
TrajectoryCrafter & 0.477 & 0.179 & 0.284 & 0.138 & 0.099 & 0.071 & 0.156 & 0.377 \\
ReCamMaster       & 0.495 & 0.075 & 0.303 & 0.037 & 0.039 & 0.010 & 0.093 & 0.056 \\
CamCloneMaster    & 0.477 & 0.064 & 0.220 & 0.028 & 0.196 & 0.058 & 0.220 & 0.111 \\
Vista4D           & \textbf{0.771} & 0.587 & 0.303 & 0.156 & 0.472 & 0.288 & 0.519 & 0.528 \\
\midrule
\textbf{Ours}     & 0.743 & \textbf{0.761} & \textbf{0.817} & \textbf{0.743} & \textbf{0.575} & \textbf{0.500} & \textbf{0.761} & \textbf{0.587} \\
\bottomrule
\end{tabular}
\end{table*}

\paragraph{User study.}
\label{sec:user_study}
We conduct a blind pairwise study with 55 participants and 16 video pairs: nine comparisons against Vista4D and seven against CamCloneMaster. Each comparison includes the reference video, the requested behavior, and two anonymized outputs. Participants assess (1) camera-control alignment and (2) visual quality and fidelity to the reference scene, with an option to rate the outputs equally. The instructions and interface are provided in Appendix~\ref{app:user_study}.

\begin{table*}[t]
\centering
\caption{\textbf{Ablation of design choices} on a subset of 33 scenes. Metrics and aggregation follow Tab.~\ref{tab:consolidated_updated}. Appearance scores should be interpreted alongside motion success, as outputs with little camera movement can retain high reference similarity. Best values in each column are shown in \textbf{bold}. Layer-selection and rank ablations appear in Tab.~\ref{tab:ablation_full}.}
\label{tab:ablation_main}
\begin{tabular}{lcccccc}
\toprule
& \multicolumn{2}{c}{Motion} & \multicolumn{2}{c}{Subject appearance} & \multicolumn{2}{c}{VLM judge} \\
\cmidrule(lr){2-3}\cmidrule(lr){4-5}\cmidrule(lr){6-7}
Variant & Sem.\ motion~$\uparrow$ & Success~$\uparrow$ & DINO~$\uparrow$ & ID-Sim~$\downarrow$ & Sem.\ motion~$\uparrow$ & Identity~$\uparrow$ \\
\midrule
\textbf{Full (Ours)}          & \textbf{0.654} & \textbf{0.706} & 0.636 & 0.283 & \textbf{0.610} & 0.737 \\
\quad$-$ background loss      & 0.611 & 0.634 & 0.638 & 0.284 & 0.602 & 0.720 \\
\quad$-$ motion modulation    & 0.530 & 0.554 & 0.662 & 0.259 & 0.551 & 0.768 \\
Shared LoRA ($AB$)            & 0.306 & 0.277 & 0.704 & 0.225 & 0.436 & \textbf{0.809} \\
Per-motion LoRA ($A_iB_i$)    & 0.455 & 0.447 & 0.642 & 0.276 & 0.540 & 0.742 \\
Text-conditioned              & 0.245 & 0.228 & \textbf{0.713} & \textbf{0.213} & 0.396 & 0.805 \\
\bottomrule
\end{tabular}
\end{table*}

\paragraph{VLM evaluation.}
We use Claude Opus~4.8~\cite{anthropic2026opus} as a complementary judge, without revealing method identities. In the first pass, the judge receives 16 time-stamped output frames and assesses events required by the requested behavior, such as an orbit around the subject or a final frontal view. In the second pass, it compares reference and output montages to assess appearance preservation, accounting for attributes hidden by viewpoint changes.

\paragraph{Qualitative results.}
Fig.~\ref{fig:baseline_full} provides baseline comparisons and Fig.~\ref{fig:results} illustrates \paper across different scenes and camera behaviors. Each example is generated independently from its reference video and requested label.

The examples highlight challenges in adapting camera motion to the subject. The trajectory-conditioned baselines sometimes move away from the subject or fail to reach the requested view, while CamCloneMaster can transfer motion that does not maintain the intended framing in the target scene. Vista4D preserves scene appearance well in several examples but struggles with some close-ups and viewpoint changes exposing previously unobserved sides of the subject. \paper more consistently realizes the requested camera--subject relationship in these examples. Additional results appear in Fig.~\ref{fig:more_qual_appendix}, ~\ref{fig:baseline_appendix} and ~\ref{fig:baseline_appendix_cont} .

\paragraph{Quantitative results.}
Tab.~\ref{tab:consolidated_updated} reports results over all 109 scenes and eight behaviors, averaging first within each behavior and then across behaviors. \paper achieves a semantic-motion success rate of $68.6\%$, compared with $45.3\%$ for Vista4D and at most $22.3\%$ for the remaining baselines. Subject appearance is comparable to Vista4D: DINO similarity is $0.594$ versus $0.592$, and ID-Sim distance is $0.293$ versus $0.307$.

The component breakdown in Tab.~\ref{tab:motion_breakdown} indicates that baseline outputs often exhibit camera movement but do not achieve the requested subject-relative view. Their mean view scores range from $0.18$ to $0.36$, compared with $0.73$ for \paper. The largest success-rate differences occur for \emph{follow behind} and \emph{follow front}: \paper reaches approximately $74\%$ and $82\%$, compared with at most $16\%$ and $30\%$ for the baselines (Tab.~\ref{tab:motion_specific}). Vista4D performs slightly better on plain \emph{follow} ($77.1\%$ versus $74.3\%$).

The VLM evaluation shows the same overall motion ranking. \paper obtains a motion score of $0.621$, compared with $0.476$ for Vista4D, and matches its identity score of $0.750$.

Additional quantitative results are shown in Tab.~\ref{tab:motion_breakdown}.

\begin{figure*}[th!]
    \centering
    \includegraphics[width=1.0\textwidth]{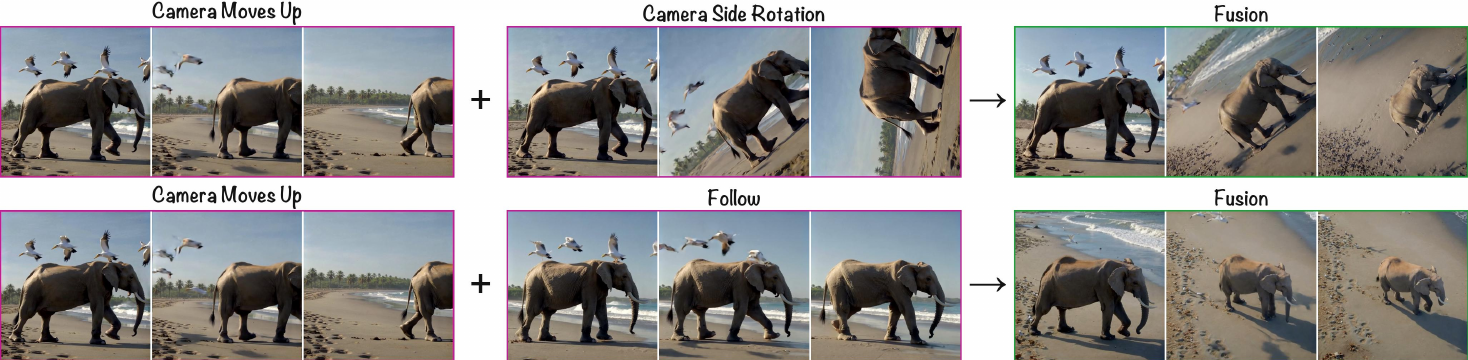}
\caption{\textbf{Motion fusion.} Weighted combinations of motion-specific LoRA factors produce composite camera behaviors without additional training. \textbf{Left:} \emph{Move up}. \textbf{Middle:} \emph{Side rotation} (top) and \emph{follow} (bottom). \textbf{Right:} The corresponding combinations, showing upward camera movement combined with rotation (top) or subject-following (bottom).}
    \label{fig:fusion}
\end{figure*}

\paragraph{Human preferences.}
As shown in Fig.~\ref{fig:user_study}, participants prefer \paper over Vista4D in $66.3\%$ of responses for camera-control alignment and $62.2\%$ for visual quality and scene fidelity. Vista4D is preferred in $17.4\%$ and $19.6\%$, respectively; the remaining responses indicate a tie. Against CamCloneMaster, \paper is preferred in $94.5\%$ of responses for camera alignment and $76.6\%$ for visual quality, compared with $3.4\%$ and $9.9\%$ for the baseline.

\subsection{Ablation Study}
\label{sec:ablation}

We evaluate the main design choices on a 33-scene subset with trackable subjects across all eight trained camera behaviors.  Tab.~\ref{tab:ablation_main} reports the main component ablations, with representative outputs shown in Fig.~\ref{fig:ablation}. Full results, including layer-selection and rank variants, are provided in Appendix~\ref{sec:ablation_full}.

\paragraph{Parameter sharing and motion conditioning.}
Independent per-motion adapters achieve a success rate of $0.447$, while a fully shared adapter reaches $0.277$, compared with $0.706$ for the full model. Removing motion-conditioned modulation reduces success to $0.554$. These results support combining shared adaptation parameters with motion-specific factors and activation conditioning.

Fig.~\ref{fig:ablation} illustrates the corresponding differences in camera behavior and content preservation. For \emph{follow top view} (left), the full model raises the camera above the alien ship and looks down onto the city. Without modulation, the camera barely departs from the input viewpoint. The fully shared adapter produces a following motion without reaching the requested top view, while the per-motion adapter also fails to achieve the elevated viewpoint. For \emph{follow front view} (right), the full model moves in front of the walking man while preserving both pedestrians and the building. The shared adapter remains beside the subjects, while the per-motion adapter generates only the man. 

Replacing the motion mechanism with text conditioning reduces success to $0.228$. This variant frequently confuses related behaviors or fails to produce the requested viewpoint change. In Fig.~\ref{fig:ablation}, it remains close to the input viewpoint for both the top-view and front-view instructions.

\paragraph{Background-consistency loss.}
Removing the background loss reduces success from $0.706$ to $0.634$. The VLM judge also assigns lower scores for artifact-free generation ($0.693$ versus $0.736$), viewpoint plausibility ($0.867$ versus $0.890$), and overall quality ($0.642$ versus $0.668$). Subject identity and scene-preservation scores remain approximately unchanged, suggesting that the main measured benefit is improved motion execution and fewer visual artifacts.

The examples in Fig.~\ref{fig:ablation} illustrate the visual impact of removing this loss. In the top-view example, the camera rises, but the city below loses its structure and becomes an abstract texture. In the front-view example, the camera reaches the requested viewpoint, but the second pedestrian disappears. Similar content changes occur without modulation, whereas the full model better preserves these scene elements.

\paragraph{Appearance and motion execution.}
Several weaker variants obtain higher appearance scores than the full model. These scores must be interpreted alongside their lower motion success: outputs that remain close to the input viewpoint face a less demanding appearance-preservation problem. For every variant, failed clips have higher DINO similarity than successful clips, with differences ranging from $0.06$ to $0.18$.

On clips where both a variant and the full model execute the requested behavior, the appearance differences generally narrow. The shared-adapter and text-conditioned variants retain small DINO advantages on their jointly successful subsets. We therefore report appearance preservation and motion execution separately rather than treating either as a complete measure of performance.

\subsection{Motion Fusion}
\label{sec:fusion}

We investigate whether camera behaviors can be combined without training on examples of their composition. We train geometric motions using data from MultiCamVideo~\cite{bai2025recammaster}, such as \emph{move up}, alongside our semantic behaviors. At inference time, we combine motion-specific factors within the shared basis:
\begin{equation}
    W' = W + (w_1A_i+w_2A_j)B,
\end{equation}
where $w_1$ and $w_2$ control the contributions of the two factors. We retain the modulation associated with one constituent motion. In the tested compositions, choosing either constituent's modulation produces qualitatively similar behavior, suggesting that the mixed LoRA factors primarily determine the resulting camera motion.

Fig.~\ref{fig:fusion} shows qualitative examples. Combining \emph{move up} with \emph{follow} produces a camera that rises while keeping the subject framed. Combining \emph{move up} with \emph{side rotation} produces simultaneous upward and rotational movement. These examples illustrate composition of learned camera behaviors without additional training or paired examples of the combined behavior.

\subsection{Limitations}
\label{sec:limitations}

\paper remains challenged by complex subject dynamics. In multi-subject scenes, the motion label does not specify which subject to follow, and the model typically selects the visually dominant subject. Abrupt changes in direction can cause the generated camera to lag behind the subject. Generalization is also constrained by the training dataset's limited size, subject diversity, and visual domain.

Our background-consistency loss provides supervision only in regions estimated to be visible in both the reference and target videos. 
Under large viewpoint changes, including some front- and rear-following views, the estimated co-visibility mask can be empty, making the auxiliary loss inactive. 
This can occur because little background is shared or because the reconstruction fails to recover existing overlap. Newly revealed regions therefore rely on the base training objective and the model's generative prior, which do not guarantee geometric consistency. In challenging cases, the model may confuse spatial relationships or reproduce reference-view elements in locations inconsistent with the new viewpoint.

Broader training data, explicit subject selection, more robust visibility estimation, and additional geometry-aware constraints are promising directions for future work.

\section{Conclusion}

We introduced \paper, a framework for semantic camera motion control that defines camera behavior through high-level intent rather than explicit trajectories or reference motions. To support this task, we constructed a dataset of paired videos covering eight semantic camera motions, together with a 109-scene benchmark and objective, subject-anchored evaluation metrics. Our method uses a parameter-efficient architecture featuring shared low-rank factorization and motion-conditioned modulation, enabling robust control and strong visual consistency.
Extensive experiments demonstrate that \paper generalizes to unseen scenes and achieves higher average semantic-motion accuracy than the evaluated baselines, including the recent video-reshooting method Vista4D, while maintaining comparable subject identity preservation.
Furthermore, qualitative results demonstrate motion-fusion capabilities within the shared low-rank space. We hope this work encourages future research into cinematic video generation driven by semantic intent.

\bibliographystyle{eg-alpha-doi}
\balance

\bibliography{references}
\clearpage
\nobalance

\appendix
\clearpage
\setcounter{figure}{0}
\renewcommand{\thefigure}{A\arabic{figure}}
\renewcommand{\theHfigure}{appendix.\arabic{figure}}

\setcounter{table}{0}
\renewcommand{\thetable}{A\arabic{table}}
\renewcommand{\theHtable}{appendix.\arabic{table}}
\section{Implementation Details}
\label{sec:impl}

\begin{figure}[!h]
\centering
\includegraphics[width=\linewidth]{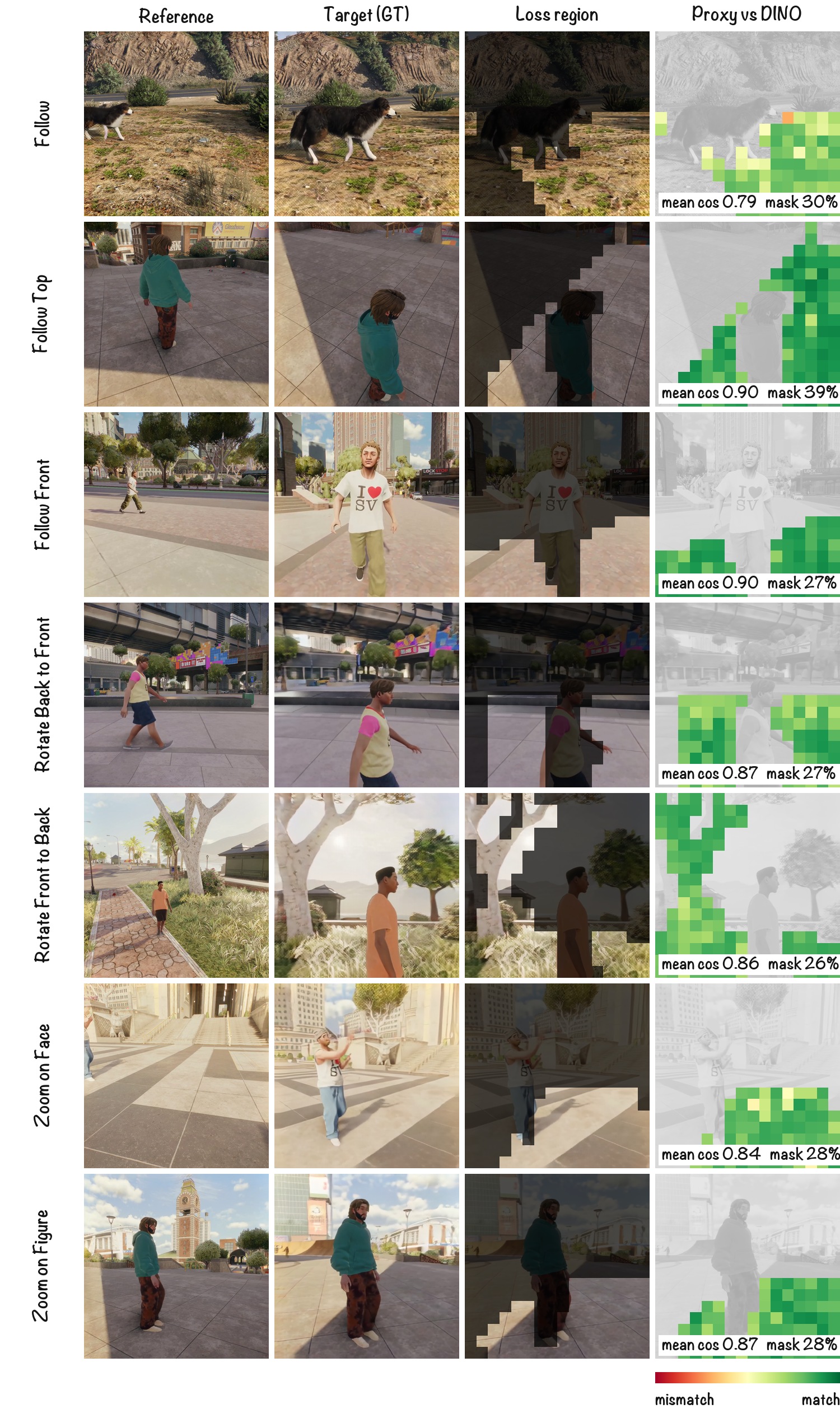}
\caption{\textbf{Where the background-consistency loss looks.} One row per semantic motion: reference and ground-truth target frames, the loss region $\mathcal{M}=\mathrm{covis}\wedge\neg\,\mathrm{subject}$, and the per-patch cosine agreement between the proxy features $g(\hat z_0)$ and DINO$(x^\star)$, shown only on $\mathcal{M}$ (green = match, red = mismatch). The mask excludes the subject and the regions the reference never saw, so the loss is applied only where the background should persist.}
\label{fig:bg_loss_where}
\end{figure}

\begin{table}[t]
\caption{Training pairs for the eight semantic camera motions.}
\label{tab:dataset_v9}
\centering
\small
\begin{tabular}{clc}
\toprule
\# & Motion type & Pairs \\
\midrule
1  & Follow                          &  97 \\
2  & Follow front view               & 103 \\
3  & Follow top view                 &  84 \\
4  & Follow behind           & 86 \\
5  & Rotate back $\rightarrow$ front & 108 \\
6  & Rotate front $\rightarrow$ back & 100 \\
7  & Rotate + zoom on face           &  41 \\
8  & Rotate + zoom on figure         &  42 \\
\midrule
   & \textbf{Total}                  & \textbf{661} \\
\bottomrule
\end{tabular}
\end{table}

\paragraph{Training.}
Our method builds on LTX-2~\cite{hacohen2025ltx2}, a video diffusion transformer trained with flow matching. We train our adaptation in the low-resolution stage, concatenating reference-video latent tokens with noisy target-video tokens so that the model learns the requested transformation conditioned on the source video. We jointly train all eight semantic camera behaviors on 661 video pairs for 50,000 steps using a single NVIDIA H100 80GB GPU. We use a batch size of 1, a learning rate of $1\times10^{-4}$, and a LoRA rank of $r=64$. Training videos are center-cropped and resized to $512\times512$ pixels and contain 81 frames.

\begin{table*}[t]
\centering
\caption{\textbf{Semantic-motion score per motion type}: $M$ is the minimum of the listed components.}
\label{tab:motion_components}
\small
\begin{tabular}{p{2.6cm}p{3.9cm}p{5.6cm}}
\toprule
Motion & Components ($M=\min$) & What they check \\
\midrule
follow & Travel, Dist.\ kept & camera moves with the subject at a constant distance \\
follow top & Travel, Dist.\ kept, Elevation & as follow, from above the subject \\
follow front & Travel, Dist.\ kept, View (front) & as follow, facing the subject's front throughout \\
follow behind & Travel, Dist.\ kept, View (back) & as follow, behind the subject throughout \\
rotate back$\to$front & View (back$\to$front), Dist.\ kept & orbits from behind to the front at a constant radius \\
rotate front$\to$back & View (front$\to$back), Dist.\ kept & orbits from the front to behind at a constant radius \\
zoom face & Zoom, Face at end & approaches or zooms until a face close-up \\
zoom figure & Zoom, Framing & approaches or zooms onto the full figure, kept in frame \\
\bottomrule
\end{tabular}
\end{table*}

\begin{table*}[t]
\centering
\caption{\textbf{Breakdown of the semantic-motion score} on the 109 scenes; each component is averaged over the motions that use it (Tab.~\ref{tab:motion_components}).}
\label{tab:motion_breakdown}
\small
\begin{tabular}{lccccccc}
\toprule
Method & Travel & Dist.\ kept & Elevation & View & Zoom & Face at end & Framing \\
\midrule
TrajectoryCrafter & 0.611 & 0.624 & 0.290 & 0.256& 0.384 & 0.530 & 0.681 \\
ReCamMaster       & 0.599 & 0.481 & 0.166 & 0.180 & 0.132 & 0.233 & 0.115 \\
CamCloneMaster    & 0.647 & 0.546 & 0.170 & 0.232 & 0.300 & 0.582 & 0.493 \\
Vista4D           & 0.703 & 0.756 & 0.716 & 0.356 & 0.595 & 0.719 & \textbf{0.909} \\
\midrule
\textbf{Ours}     & \textbf{0.843} & \textbf{0.800} & \textbf{0.824} & \textbf{0.726} & \textbf{0.683} & \textbf{0.823} & 0.826 \\
\bottomrule
\end{tabular}
\end{table*}

\paragraph{Background-consistency loss.}
We extract target features using DINOv2 ViT-S/14 at an image resolution of $224\times224$, producing a $16\times16$ patch grid aligned with the spatial latent grid. A patch is marked as co-visible if more than 30\% of its pixels satisfy the estimated visibility criterion, and as belonging to the subject if more than 50\% of its pixels fall within the subject mask. The frozen latent-to-DINO feature predictor is a four-layer convolutional network with hidden width 512. We weight the background loss by
\[
w(\sigma)=\mathrm{clip}\left(\frac{0.9-\sigma}{0.2},\,0,\,1\right),
\]
where $\sigma$ denotes the noise level. The auxiliary loss is inactive for $\sigma\geq0.9$, when the denoised estimate is unreliable, and reaches full weight for $\sigma\leq0.7$.

\paragraph{Two-stage inference.}
We follow the LTX-2 pipeline, first generating low-resolution video latents and then upsampling and refining them at higher resolution. The first stage produces videos at $512\times512$ resolution using 20--40 denoising steps. The second stage uses the pipeline's latent upsampler and distilled refinement model. We omit the highest noise level from the default refinement schedule $[0.909,\,0.725,\,0.42,\,0]$ and use $[0.725,\,0.42,\,0]$, as the higher initial noise reduced appearance consistency with the reference in our experiments. We apply our learned adaptation parameters and concatenate reference-video tokens during both stages.

\paragraph{Text conditioning.}
The motion label specifies the requested camera behavior, while the text prompt describes the scene content. In our experiments, descriptive scene prompts combined with negative prompts helped preserve the source appearance and reduce unintended content changes.

\begin{figure*}[t]
    \centering
    {\setlength{\tabcolsep}{2pt}\renewcommand{\arraystretch}{1.0}%
    \begin{tabular}{@{}c c @{\hskip 6pt} c@{}}
     & {\dsfont Reference Video} & {\dsfont Ground Truth Video} \\[2pt]
    \dslbl{Follow} & \dsimg{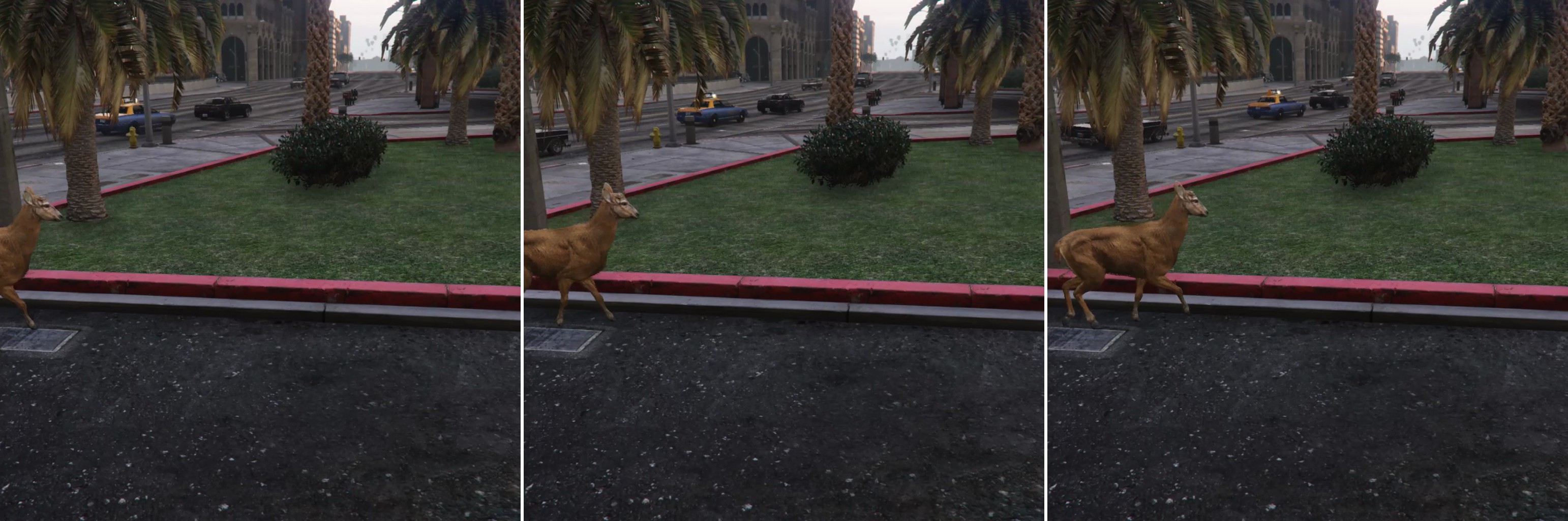} & \dsimg{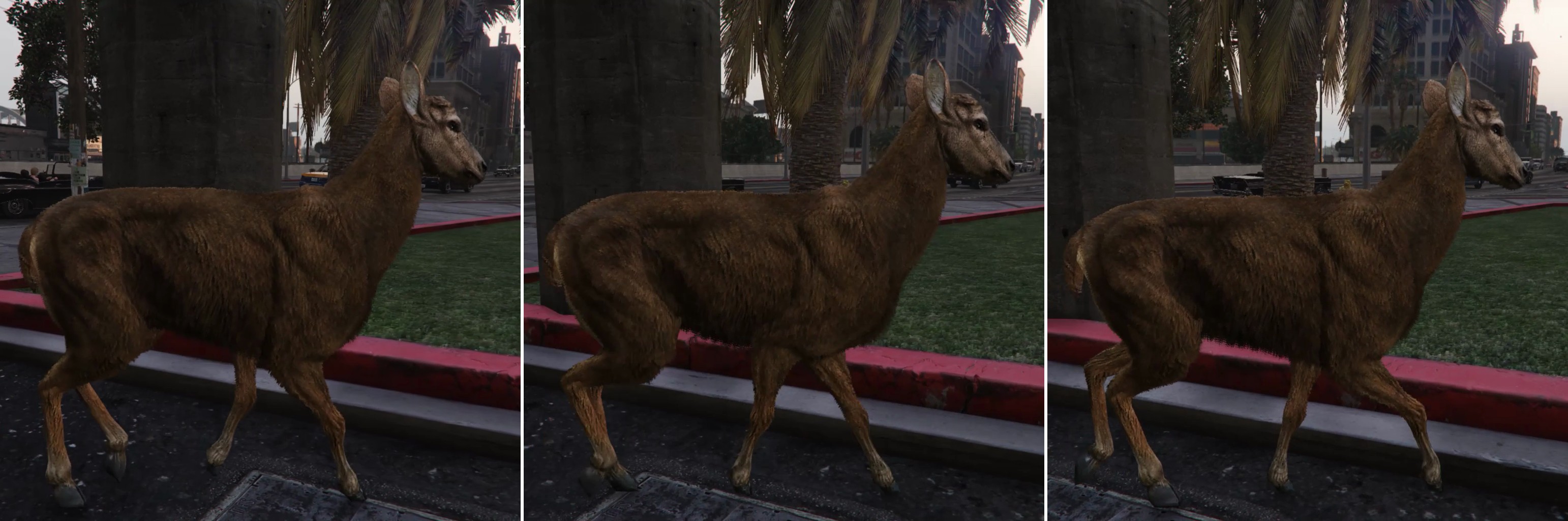} \\[3pt]
    \dslbl{Follow\\Front} & \dsimg{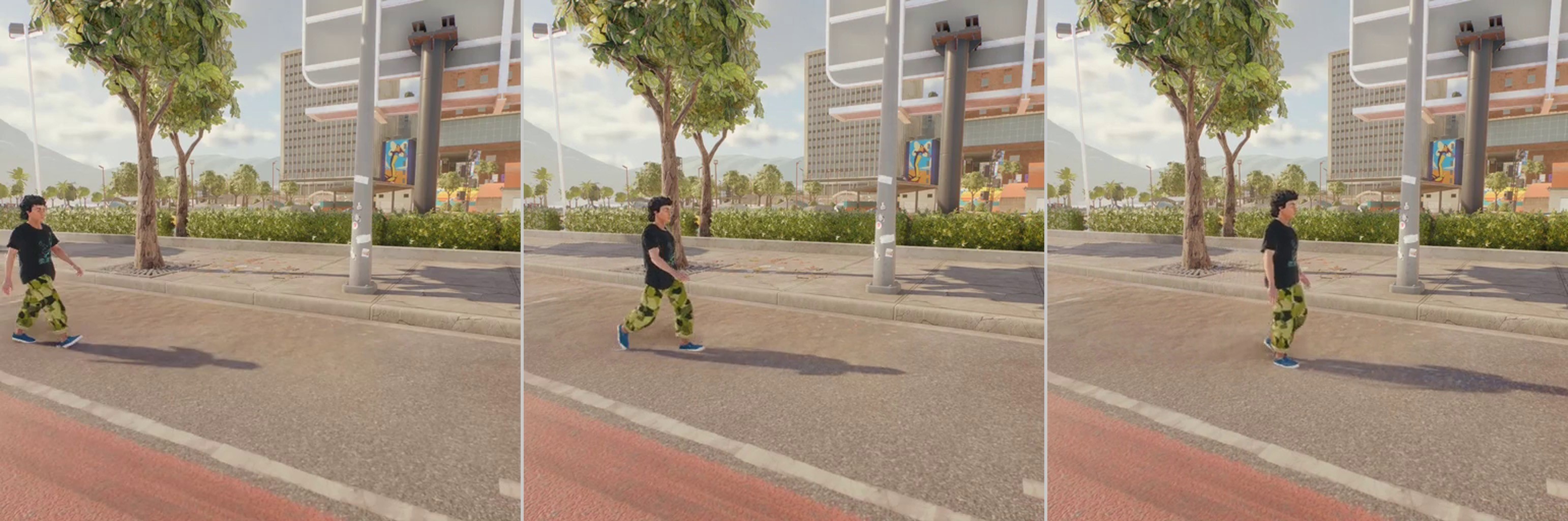} & \dsimg{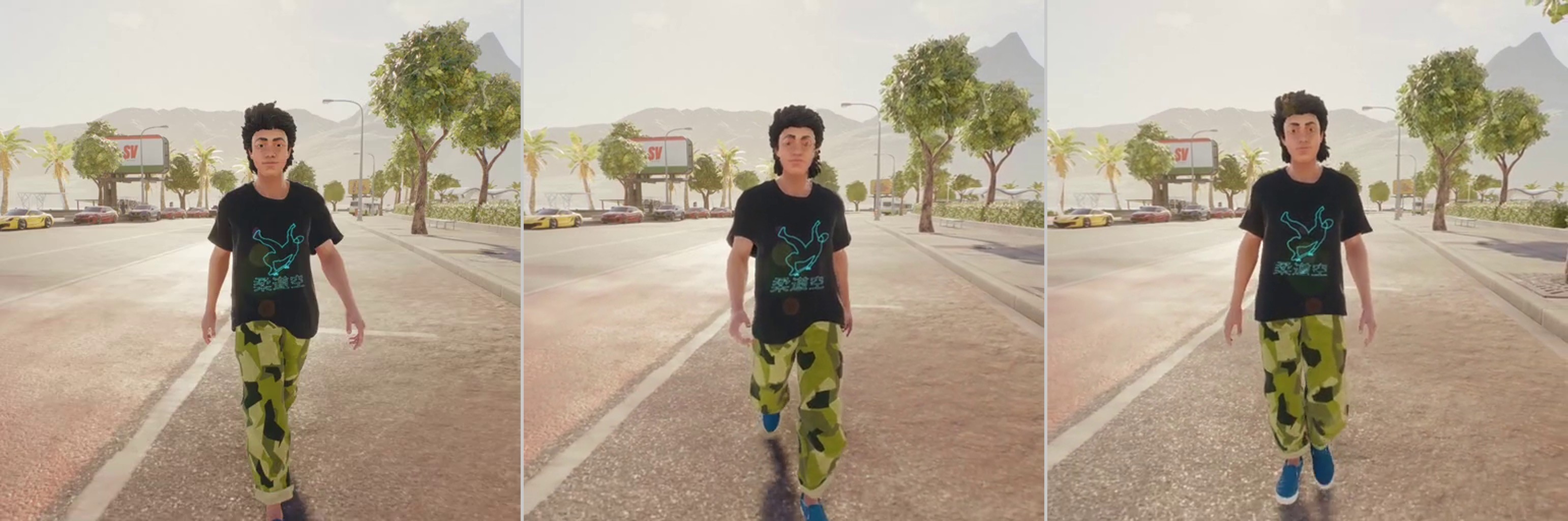} \\[3pt]
    \dslbl{Follow\\Behind} & \dsimg{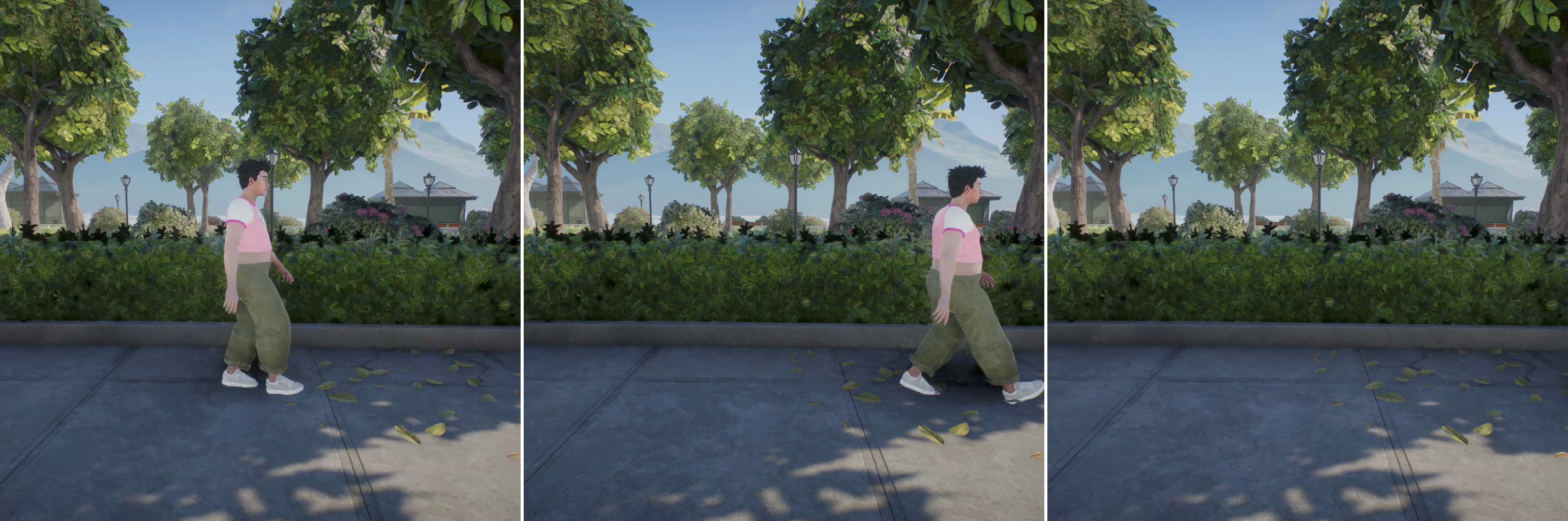} & \dsimg{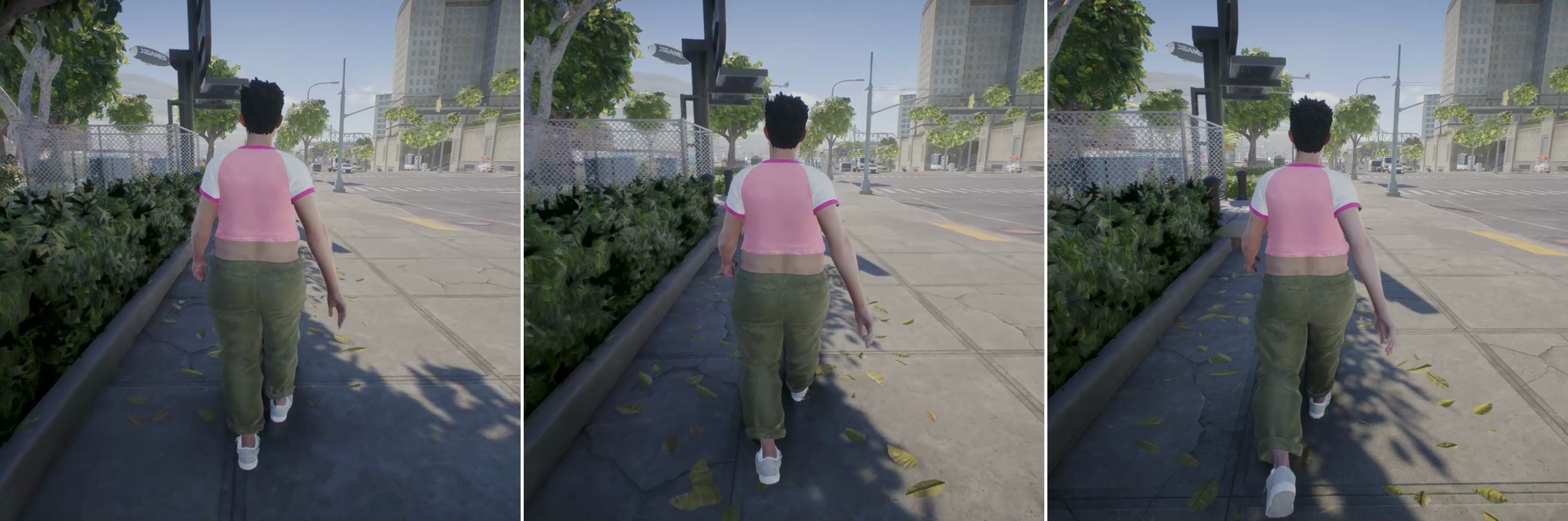} \\[3pt]
    \dslbl{Rotate\\Front $\rightarrow$ Back} & \dsimg{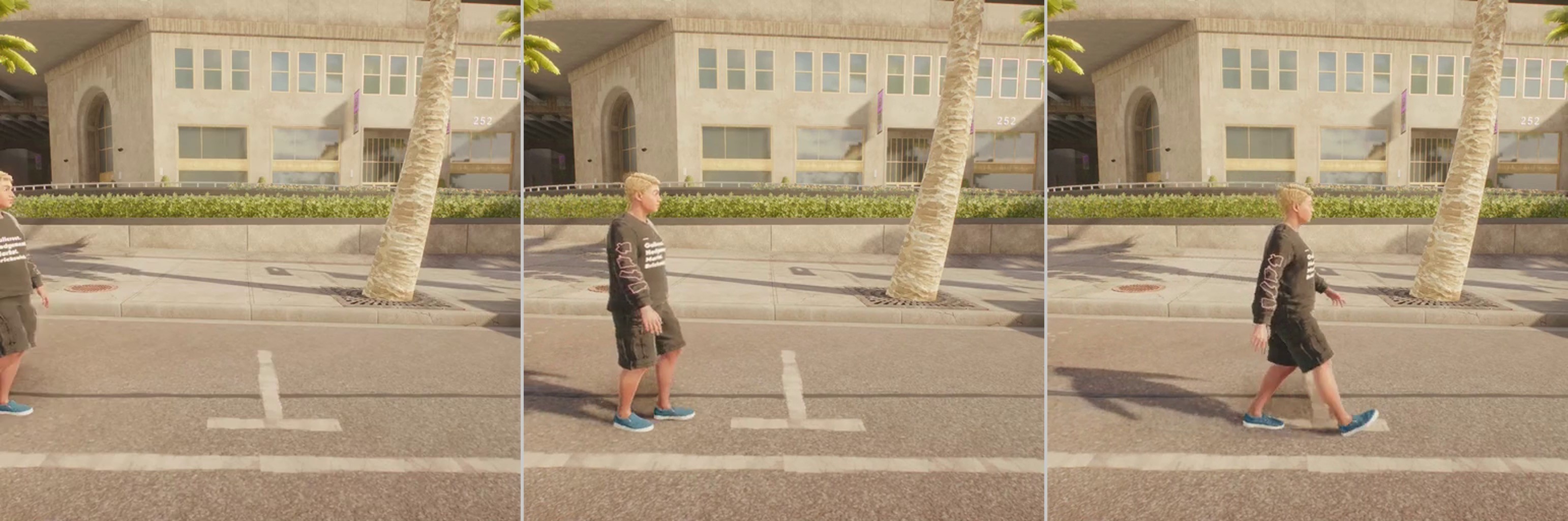} & \dsimg{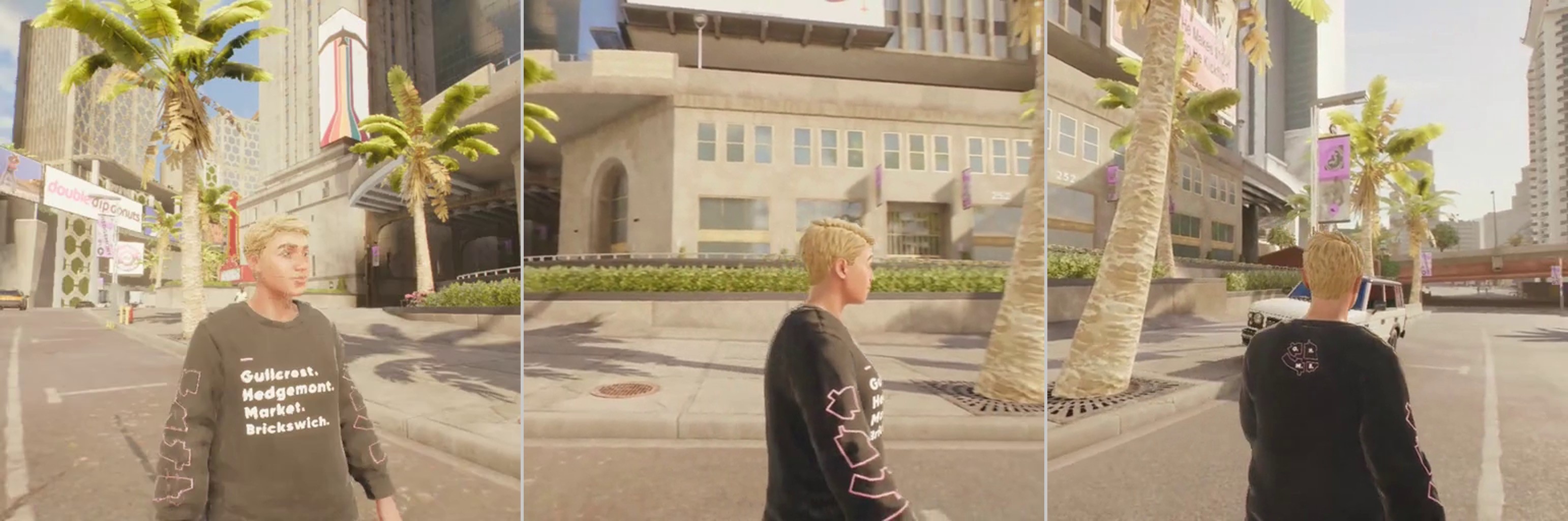} \\[3pt]
    \dslbl{Rotate $\&$ Zoom\\on Figure} & \dsimg{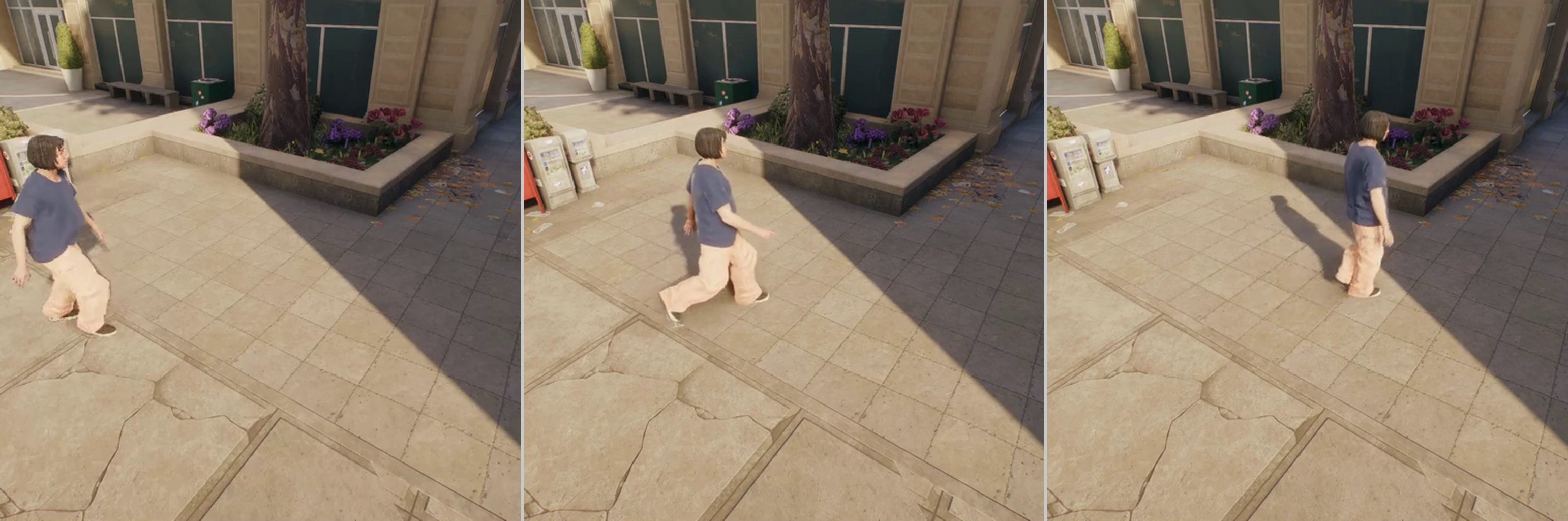} & \dsimg{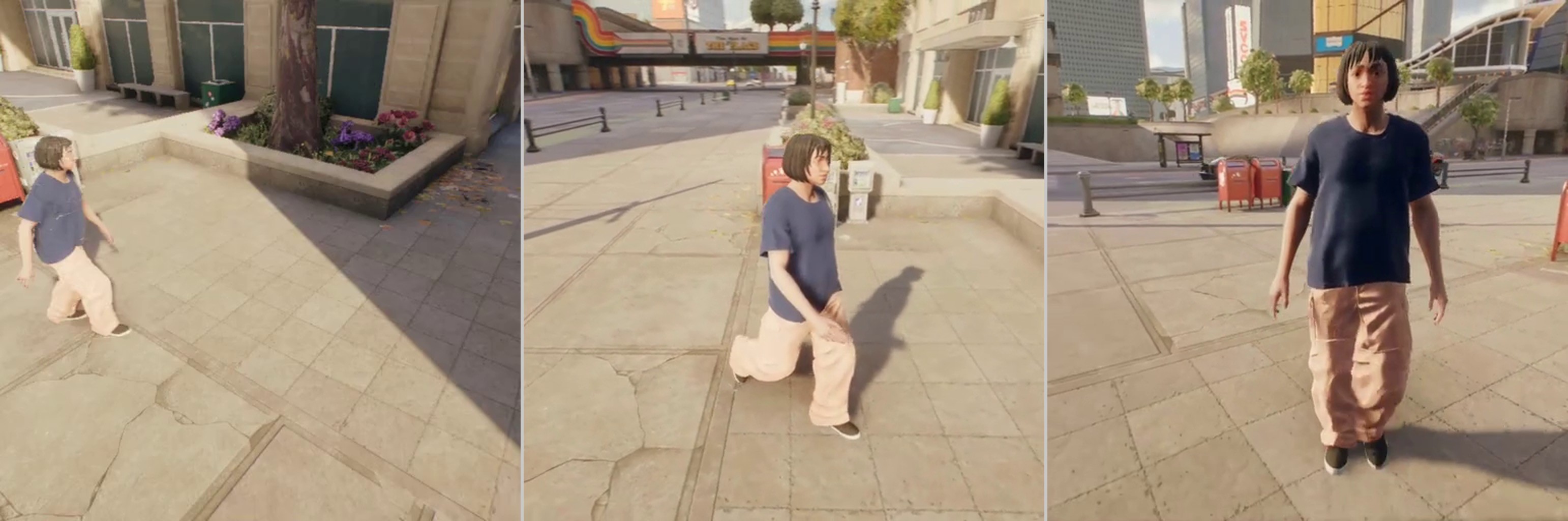} \\
    \end{tabular}}
    \caption{\textbf{Our dataset.} Examples of input--target video pairs used for training. Each pair consists of a reference video and a corresponding target video exhibiting a specific camera transformation.}
    \label{fig:dataset_samples}
    \vspace{-0.5cm}
\end{figure*}

\section{Design Choices}

\noindent \textbf{Layer Selection.} \quad
\label{sec:layer_selection}
We apply our adaptation only to the later transformer blocks. Early blocks are kept frozen to preserve low-level visual representations learned by the pretrained model, while later blocks are adapted to encode higher-level camera transformations. In our experiments, we apply LoRA to blocks indices 16--47 of the 48-block transformer.

\begin{figure*}[!h]
    \centering
    \rgrowA{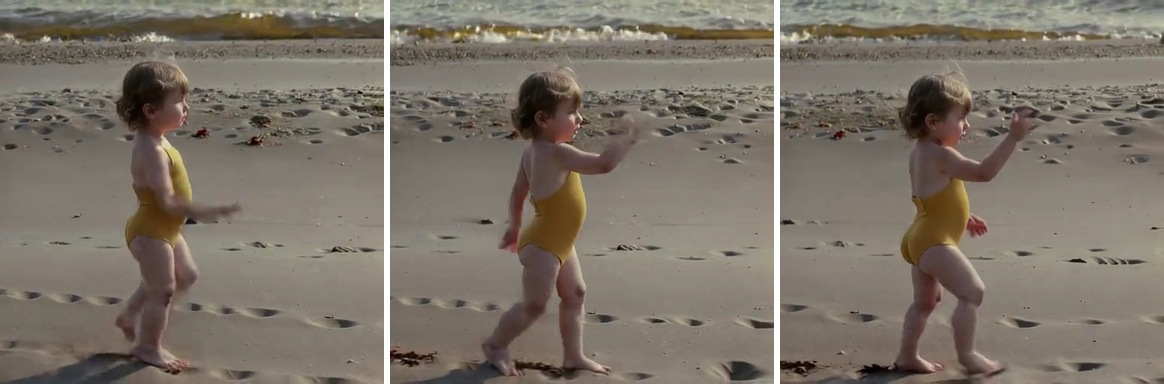}\\[\rggap]
    \rgrowB{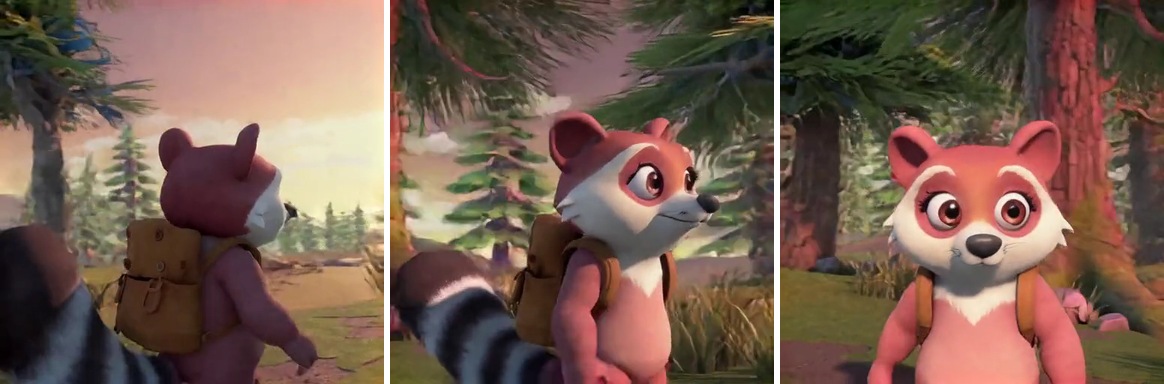}\\[\rggap]
    \rgrowA{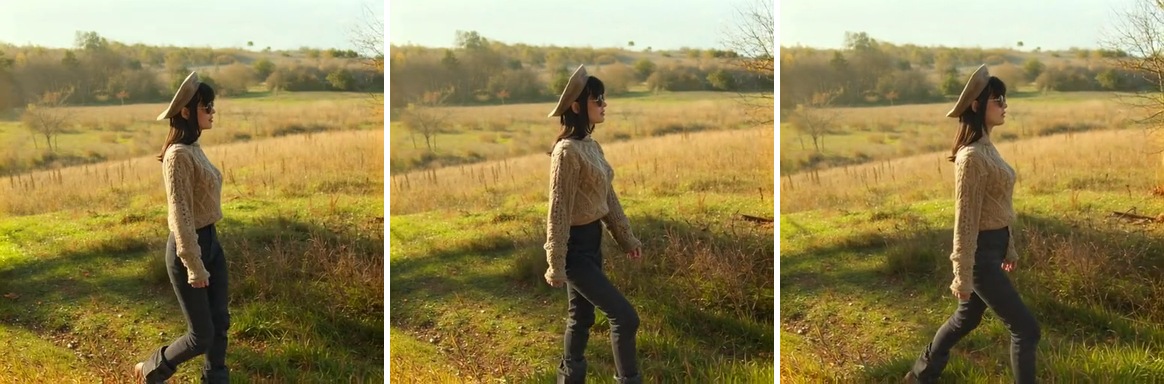}\\[\rggap]
    \rgrowB{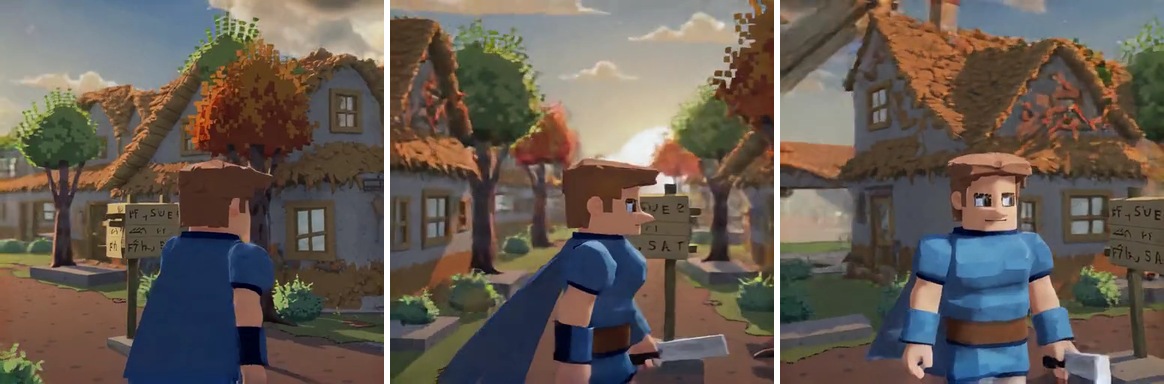}
      \caption{\textbf{Extended qualitative results.} Additional \paper generations (each cell: input video top row, \paper output bottom row; three frames over time) across further scenes and camera motions. As in Fig.~\ref{fig:results}, each clip is an independent video-to-video generation from the input plus a single motion token.}
    \label{fig:more_qual_appendix}
    \vspace{-0.5cm}
\end{figure*}

\paragraph{Co-visibility mask and its coverage.}
\label{sec:bg_loss_appendix}
We estimate the background co-visibility mask $\mathcal{M}=\mathrm{covis}\wedge\neg\,\mathrm{subject}$ by jointly reconstructing the reference and target views using STream3R~\cite{stream3r2025}. Target pixels are considered potentially co-visible if they reproject within the reference frame and have reconstruction confidence greater than 2. We aggregate these estimates into patch-level masks and exclude dynamic-subject patches identified by an instance segmenter. Because we do not perform an explicit occlusion test, the mask approximates shared visibility rather than guaranteeing it. Fig.~\ref{fig:bg_loss_where} visualizes $\mathcal{M}$ and the per-patch cosine similarity between the predicted features $h(\hat z_0)$ and the target DINOv2 features $\phi(V^\star)$.

Large viewpoint changes can reduce reconstruction confidence, causing the mask to miss genuinely shared background regions. In some front- and rear-following clips, confidence falls below the threshold throughout the frame, yielding an empty mask and disabling the auxiliary loss despite visible background overlap. In these cases, the loss provides no additional background supervision. More robust correspondence estimation could improve coverage under large viewpoint changes, while explicit occlusion handling could improve the reliability of the selected regions.

\begin{figure*}[p]
    \centering
    \includegraphics[width=0.4\textwidth]{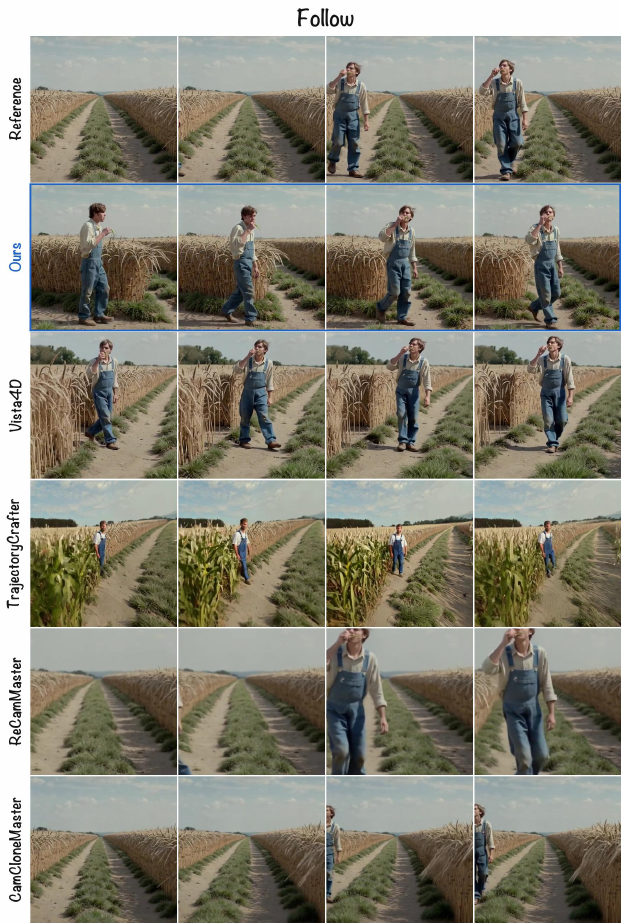}
    \includegraphics[width=0.4\textwidth]{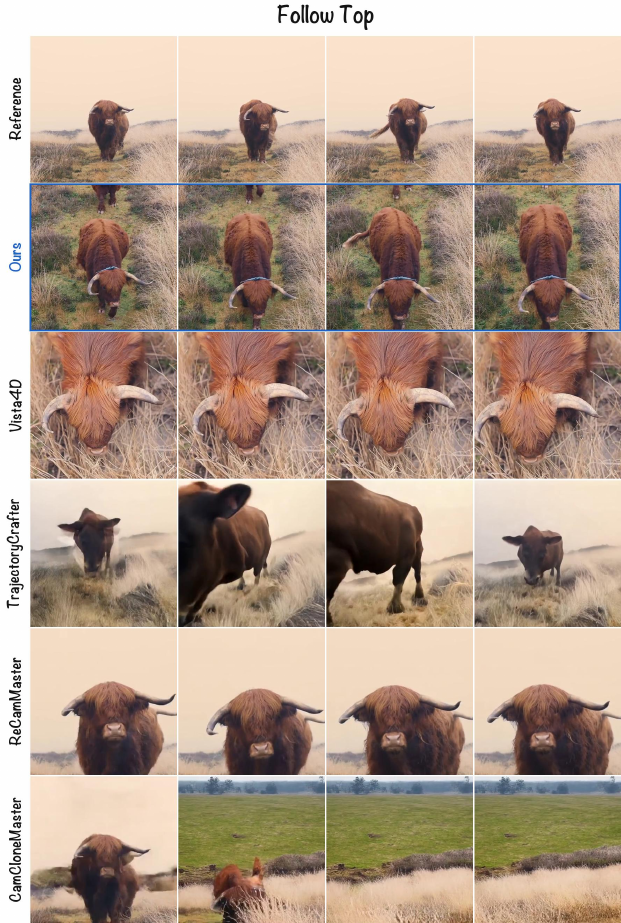}\\[4pt]
    \includegraphics[width=0.4\textwidth]{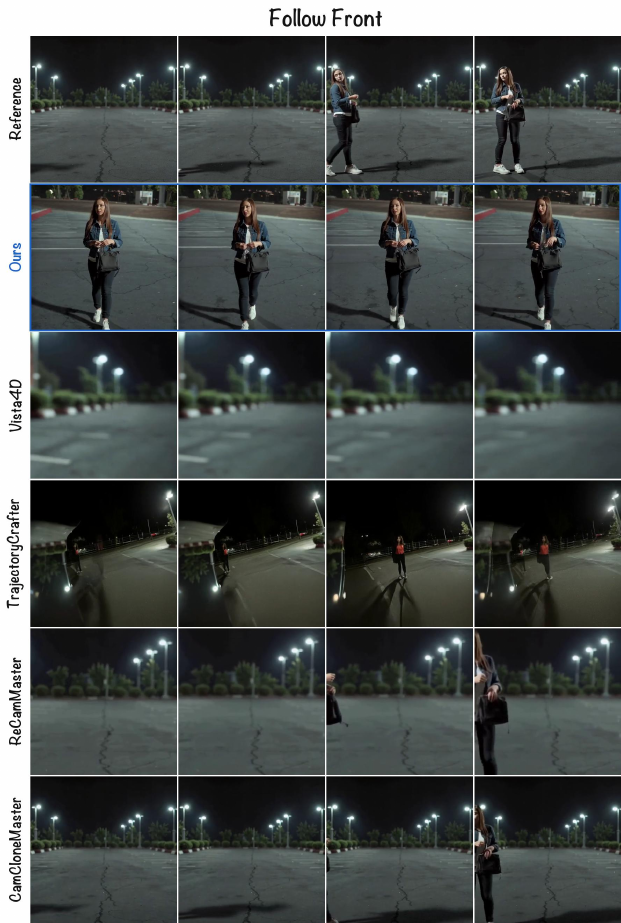}
    \includegraphics[width=0.4\textwidth]{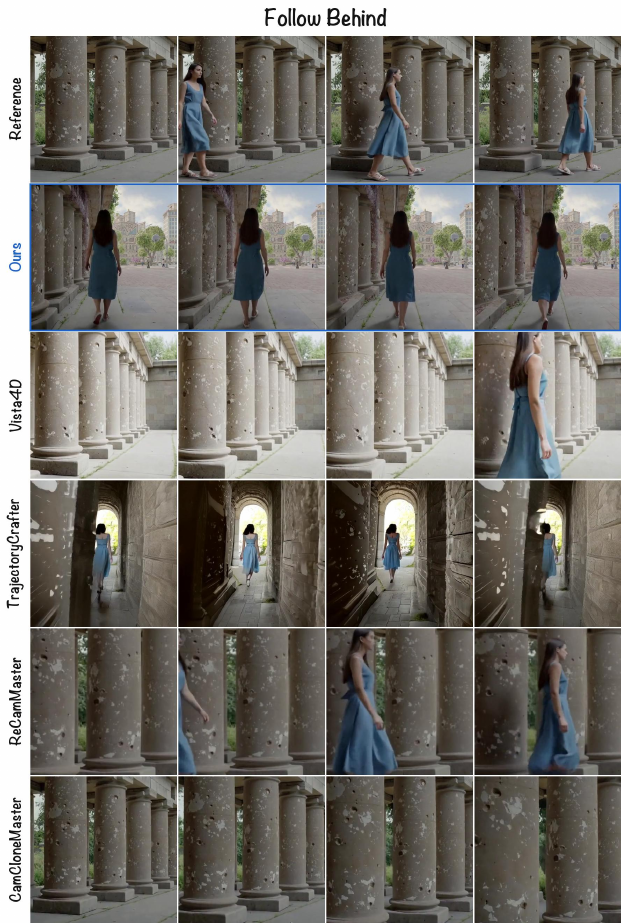}
\caption{\textbf{Additional qualitative comparisons with camera-control baselines.} \paper (highlighted in blue) against Vista4D, TrajectoryCrafter, ReCamMaster, and CamCloneMaster on further scenes, one block per camera behavior; rows show the input and method outputs, columns four frames sampled over time.}
    \label{fig:baseline_appendix}
\end{figure*}

\begin{figure*}[p]
    \centering
    \includegraphics[width=0.4\textwidth]{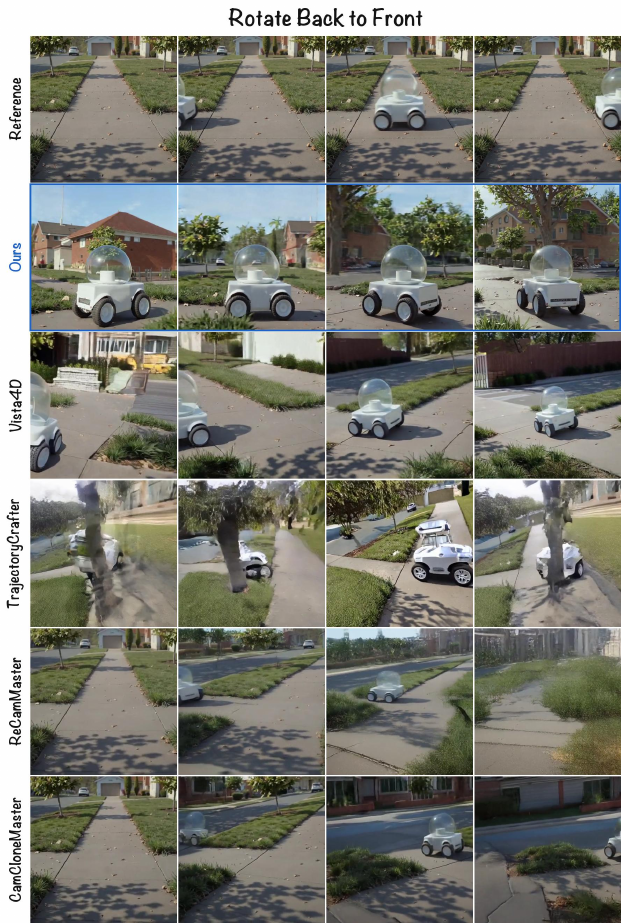}
    \includegraphics[width=0.4\textwidth]{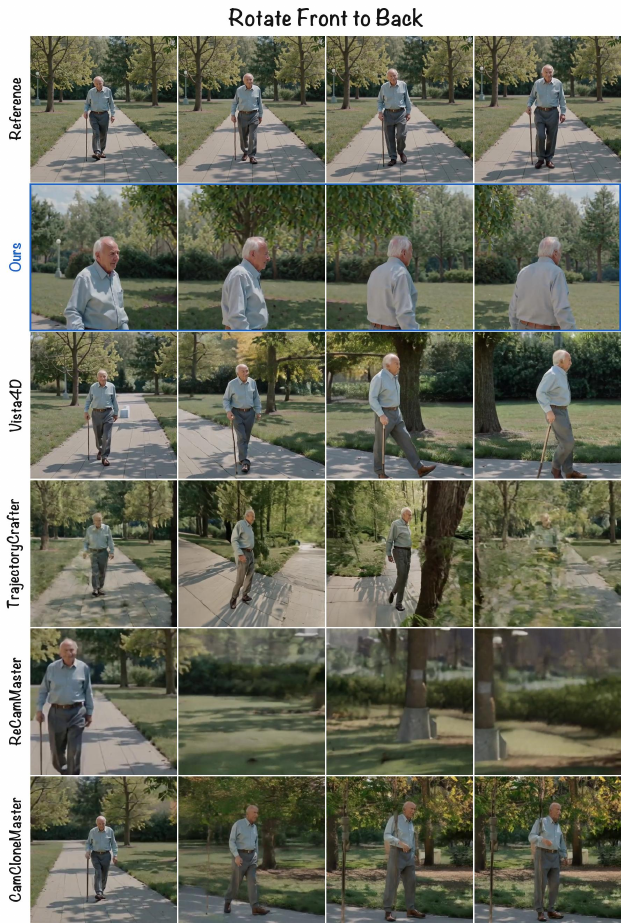}\\[4pt]
    \includegraphics[width=0.4\textwidth]{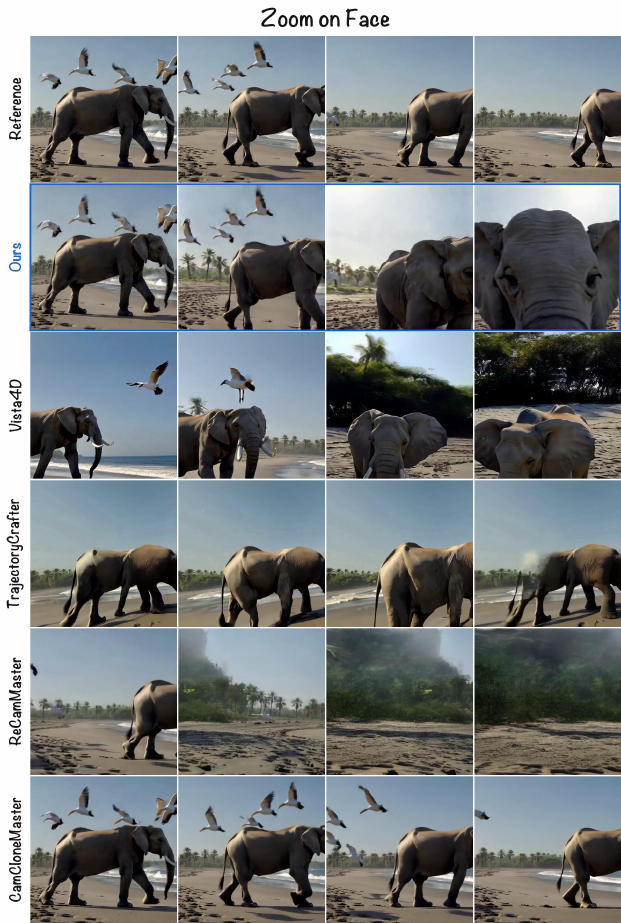}
    \includegraphics[width=0.4\textwidth]{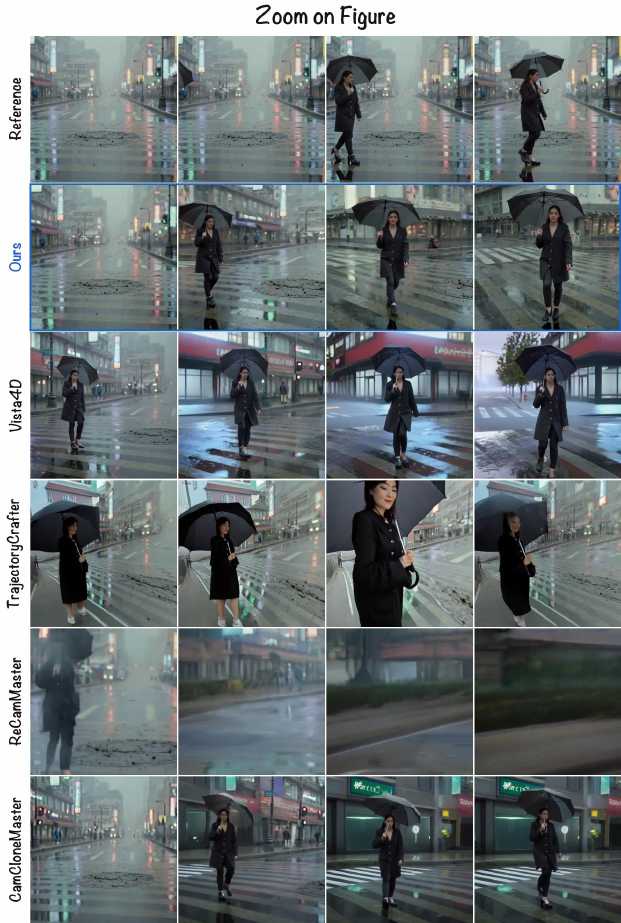}
\caption{\textbf{Additional qualitative comparisons with camera-control baselines (continued).} \paper (highlighted in blue) against Vista4D, TrajectoryCrafter, ReCamMaster, and CamCloneMaster on further scenes, one block per camera behavior; rows show the input and method outputs, columns four frames sampled over time.}
    \label{fig:baseline_appendix_cont}
\end{figure*}

\section{Output formats and evaluation protocol.}
\label{sec:baseline_protocol}
We evaluate each method at the resolution and aspect ratio used by its inference pipeline: $1024\times1024$ for \paper, $512\times512$ for TrajectoryCrafter, $832\times480$ for CamCloneMaster and ReCamMaster, and $672\times384$ for Vista4D. We retain the complete output frame rather than imposing a common square crop, which could remove the subject or alter the generated framing. Motion metrics use subject-relative geometric measurements, and appearance metrics use subject crops. The VLM judge receives letterboxed frames that preserve the full field of view.

Output lengths also differ: Vista4D produces 49 frames, TrajectoryCrafter 80, and \paper, CamCloneMaster, and ReCamMaster 81. We evaluate each method over its complete output clip without temporally resampling the generated video. As described below, the automatic motion evaluator samples 25 frames per clip, while the VLM judge receives 16 frames. This preserves each method's output sequence, although differences in clip duration and temporal sampling remain factors in the comparison.

\section{User Study: Instructions Provided to Participants}
\label{app:user_study}
\begin{figure}[!h]
  \centering
  \includegraphics[width=\linewidth]{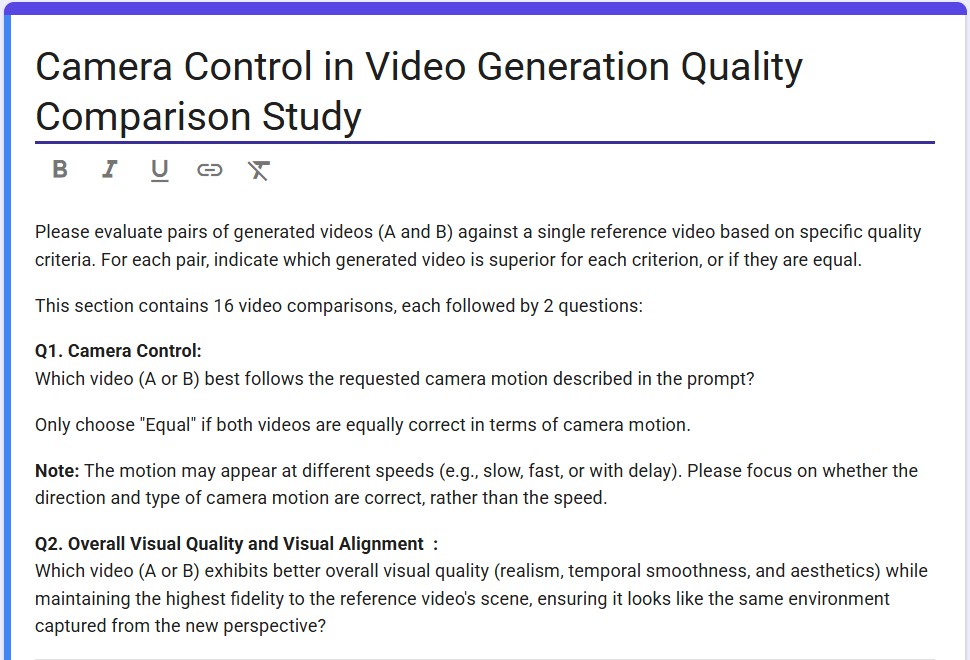}
  \caption{Screenshot of the Google Forms interface shown to participants in the user study.}
      \label{fig:user_study_interface}
\end{figure}

\begin{figure}[!h]
  \centering
  \includegraphics[width=\linewidth]{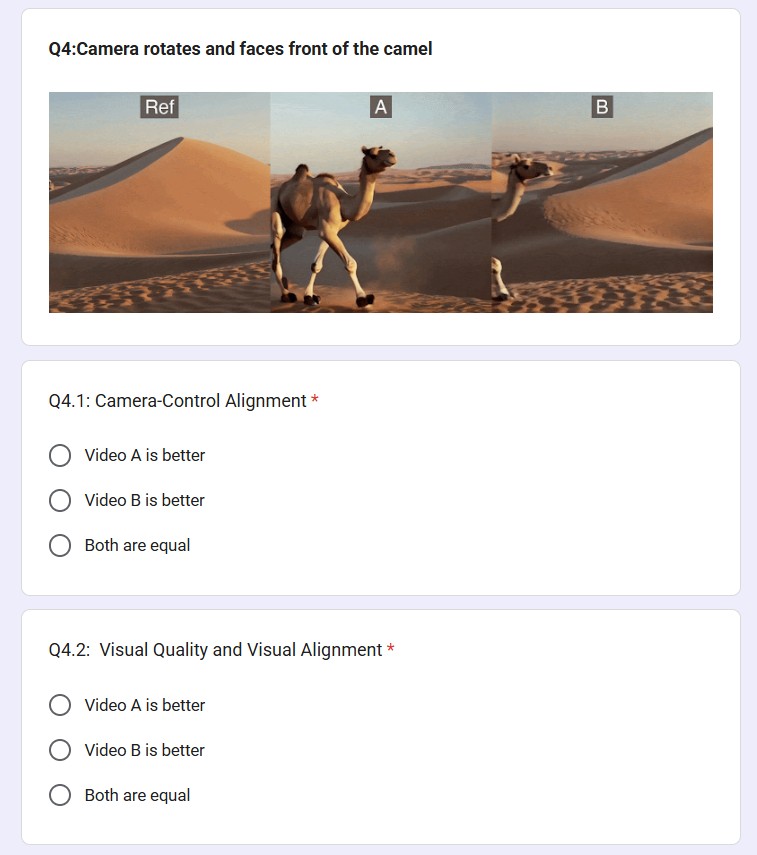}
  \caption{Example of a user study question showing intended camera motion, reference video (Ref), generated video options (A and B), and the corresponding evaluation criteria (Camera-Control Alignment and Visual Alignment).}
      \label{fig:user_study_example}
\end{figure}

As part of the evaluations we performed on our method, we conducted a user study. The study was designed to assess human preferences on generated videos comparing our method against two baselines: Vista4D and CamCloneMaster. A total of 55 participants were shown 16 video pairs in total: nine of them compared our method against Vista4D and the remaining seven against CamCloneMaster. Participants were asked to evaluate each video pair based on two criteria: \emph{camera-control alignment} (does the output camera motion match the intended type?) and \emph{visual quality \& alignment} (is the generated video visually coherent and faithful to the input scene?). The instructions provided to participants are reproduced below, and a screenshot of the interface is shown in Fig.~\ref{fig:user_study_interface}. Additionally, an example of the evaluation questions and video comparison layout presented to the participants is illustrated in Fig.~\ref{fig:user_study_example}.

\begin{quote}
\textit{Please evaluate pairs of generated videos (A and B) against a single reference video based on specific quality criteria. For each pair, indicate which generated video is superior for each criterion, or if they are equal.}

\textit{This section contains 16 video comparisons, each followed by 2 questions:}

\textbf{Q1. Camera Control:}
\textit{Which video (A or B) best follows the requested camera motion described in the prompt? Only choose ``Equal'' if both videos are equally correct in terms of camera motion. Note: The motion may appear at different speeds. Please focus on whether the direction and type of camera motion are correct, rather than the speed.}

\textbf{Q2. Overall Visual Quality and Visual Alignment:}
\textit{Which video (A or B) exhibits better overall visual quality (realism, temporal smoothness, and aesthetics) while maintaining the highest fidelity to the reference video's scene, ensuring it looks like the same environment captured from the new perspective?}
\end{quote}

\begin{table*}[!h]
\centering
\caption{\textbf{Full ablation} on the 33-scene subset. Best per column over all variants in \textbf{bold}.}
\label{tab:ablation_full}
\begin{tabular}{lcccccc}
\toprule
& \multicolumn{2}{c}{Motion} & \multicolumn{2}{c}{Subject appearance} & \multicolumn{2}{c}{VLM judge} \\
\cmidrule(lr){2-3}\cmidrule(lr){4-5}\cmidrule(lr){6-7}
Variant & Sem.\ motion~$\uparrow$ & Success~$\uparrow$ & DINO~$\uparrow$ & ID-Sim~$\downarrow$ & Sem.\ motion~$\uparrow$ & Identity~$\uparrow$ \\
\midrule
\multicolumn{7}{l}{\emph{Design components}}\\
\textbf{Full (Ours)}          & \textbf{0.654} & \textbf{0.706} & 0.636 & 0.283 & \textbf{0.610} & 0.737 \\
\quad$-$ background loss      & 0.611 & 0.634 & 0.638 & 0.284 & 0.602 & 0.720 \\
\quad$-$ motion modulation    & 0.530 & 0.554 & 0.662 & 0.259 & 0.551 & 0.768 \\
Shared LoRA ($AB$)            & 0.306 & 0.277 & 0.704 & 0.225 & 0.436 & \textbf{0.809} \\
Per-motion LoRA ($A_iB_i$)    & 0.455 & 0.447 & 0.642 & 0.276 & 0.540 & 0.742 \\
Text-conditioned              & 0.245 & 0.228 & \textbf{0.713} & \textbf{0.213} & 0.396 & 0.805 \\
\midrule
\multicolumn{7}{l}{\emph{Layer selection (ref: blocks 16--47 $=$ $-$ background loss)}}\\
blocks 0--15                  & 0.166 & 0.128 & 0.690 & 0.247 & 0.362 & 0.758 \\
blocks 16--31                 & 0.382 & 0.386 & 0.692 & 0.244 & 0.455 & 0.757 \\
blocks 32--47                 & 0.411 & 0.438 & 0.482 & 0.438 & 0.450 & 0.529 \\
blocks 0--32                  & 0.425 & 0.426 & 0.683 & 0.256 & 0.499 & 0.740 \\
blocks 0--15$+$32--47         & 0.234 & 0.212 & 0.677 & 0.257 & 0.415 & 0.764 \\
all blocks 0--47              & 0.459 & 0.470 & 0.665 & 0.273 & 0.536 & 0.734 \\
\midrule
\multicolumn{7}{l}{\emph{Rank (ref: rank 64 $=$ $-$ background loss)}}\\
rank 32                       & 0.558 & 0.580 & 0.615 & 0.312 & 0.596 & 0.724 \\
rank 128                      & 0.538 & 0.564 & 0.681 & 0.242 & 0.527 & 0.744 \\
\bottomrule
\end{tabular}
\end{table*}

\section{Full Ablation Results}
\label{sec:ablation_full}

Tab.~\ref{tab:ablation_full} gives the complete ablation on the 33-scene subset, extending the compact Tab.~\ref{tab:ablation_main} with all layer-selection (block-range) and rank variants.
Adapting blocks 16--47 with rank 64 gives the strongest motion results among the evaluated configurations, while early-block adaptation often produces weak or incorrect motion, and adaptation restricted to the latest blocks can degrade appearance, Tab.~\ref{tab:ablation_full}.

\end{document}